\documentclass[12pt,letterpaper]{article}
\usepackage[a4paper, total={7in, 10in}]{geometry}

\usepackage{helvet}
\usepackage{authblk}
\usepackage{hyperref}
\usepackage{url}            
\usepackage{booktabs}       
\usepackage{amsfonts}       
\usepackage{nicefrac}       
\usepackage{microtype}      
\usepackage{graphicx}
\usepackage{doi}
\usepackage{forest}

\usepackage{caption}
\usepackage{amsmath}
\usepackage{annotate-equations} 
\usepackage{tikz}

\usepackage{multirow}
\usepackage{colortbl}
\usepackage{tcolorbox}
\usepackage{bm}

\usepackage{pifont}       
\usepackage{bbding}       
\usepackage{fontawesome} 
\usepackage{xcolor}
\newcommand{\revise}[1]{\textcolor{black}{#1}}

\newcommand*\gpt{\raisebox{-0.16em}{\includegraphics[width=1em]{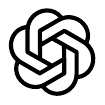}}}
\newcommand*\llama{\raisebox{-0.17em}{\includegraphics[width=1em]{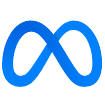}}\hspace{0.09em}}
\newcommand*\pythia{\hspace{0.03em}\raisebox{-0.09em}{\includegraphics[width=0.82em]{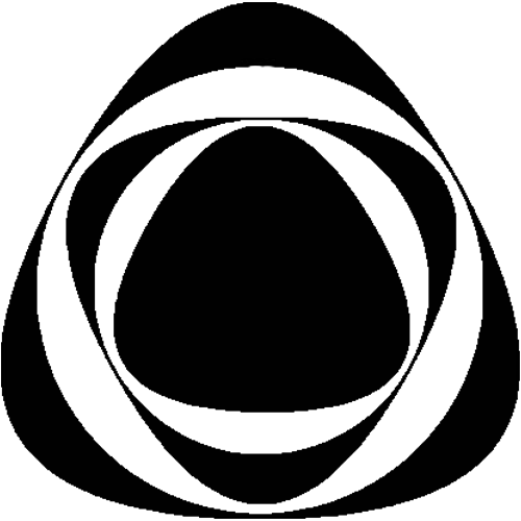}}\hspace{0.05em}}
\newcommand*\gemma{\raisebox{-0.07em}{\includegraphics[width=0.82em]{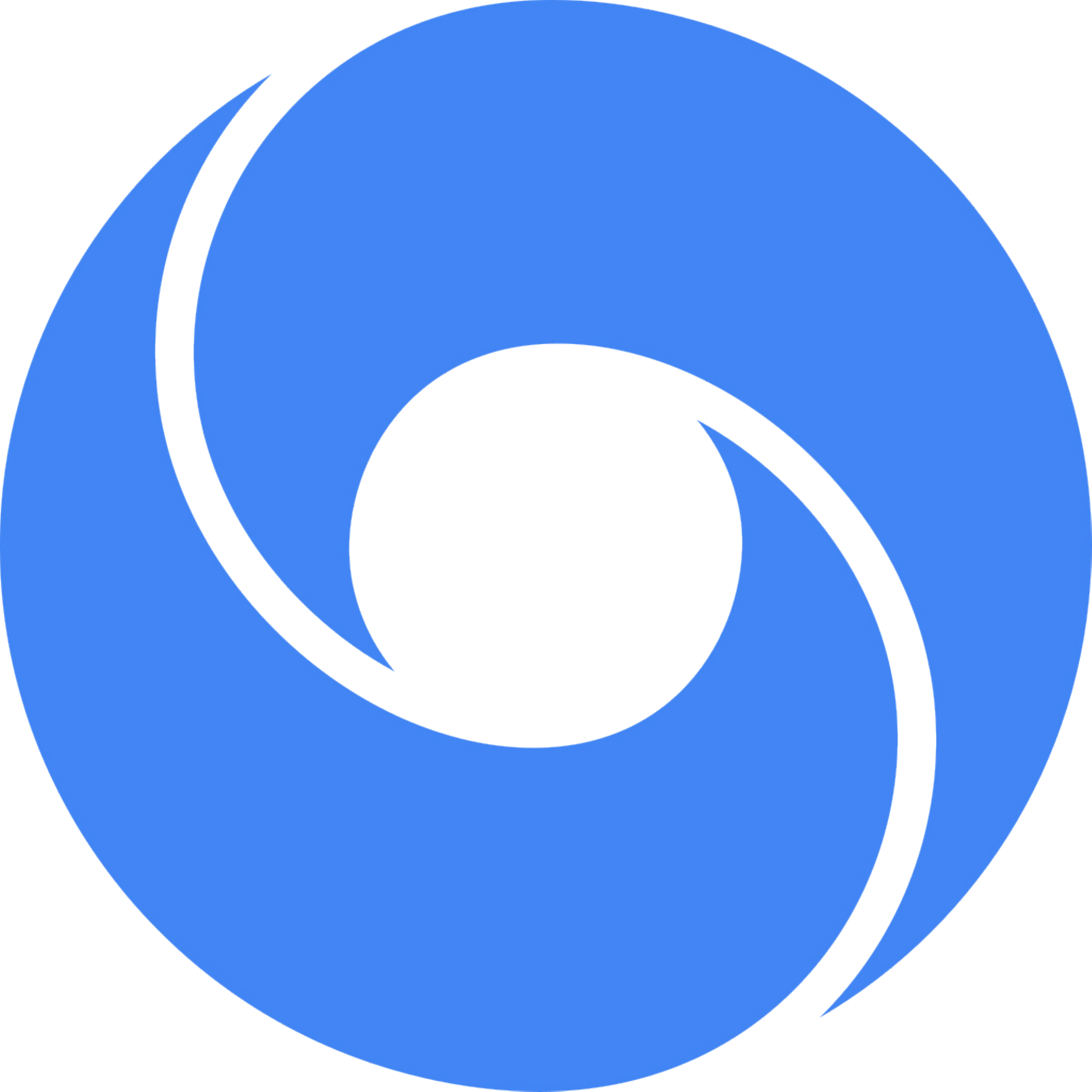}}\hspace{0.07em}}
\newcommand*\internlm{\raisebox{-0.07em}{\includegraphics[width=0.82em]{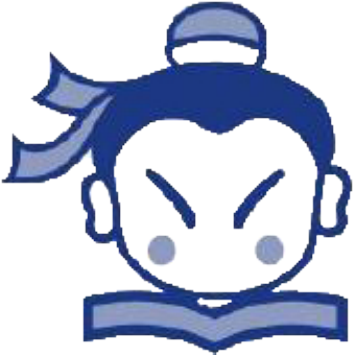}}\hspace{0.17em}}
\newcommand*\yi{\raisebox{-0.12em}{\includegraphics[width=0.82em]{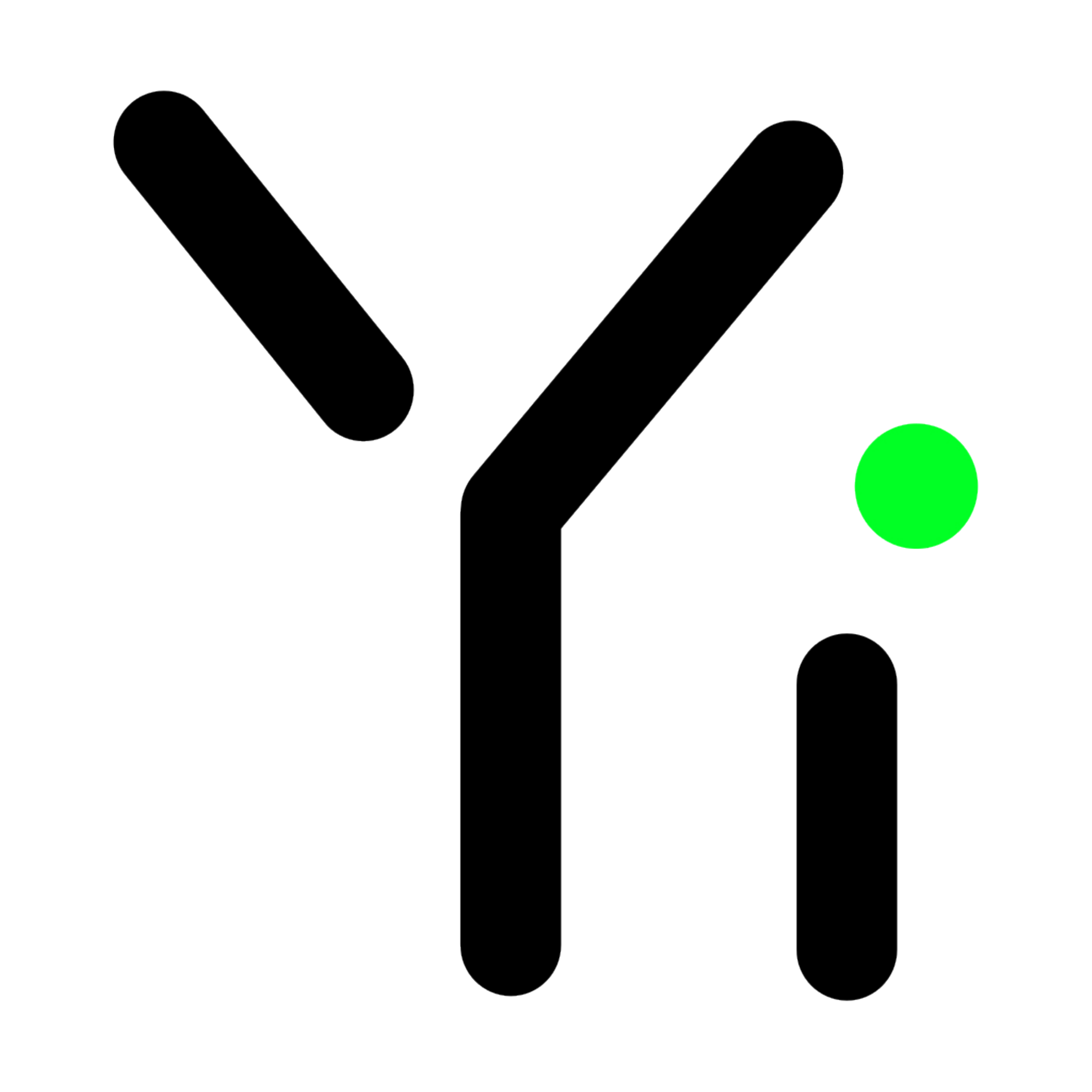}}\hspace{0.03em}}
\newcommand*\mistral{\raisebox{-0.07em}{\includegraphics[width=0.82em]{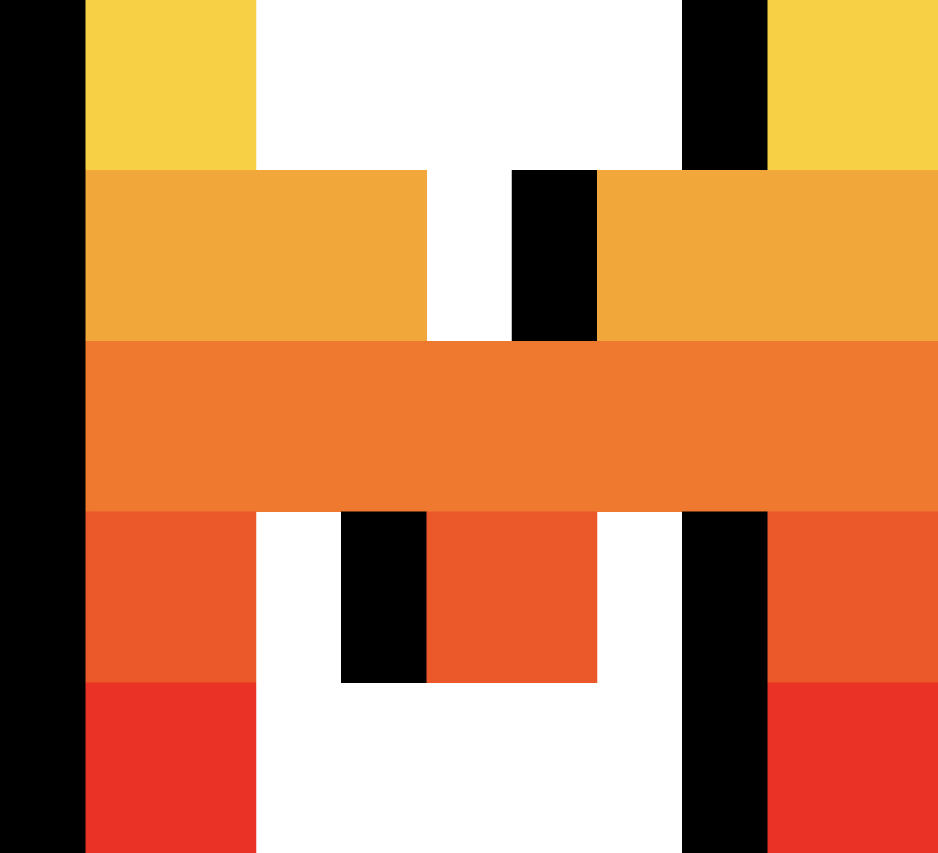}}\hspace{0.1em}}
\newcommand*\qwen{\hspace{-0.01em}\raisebox{-0.1em}{\includegraphics[width=0.82em]{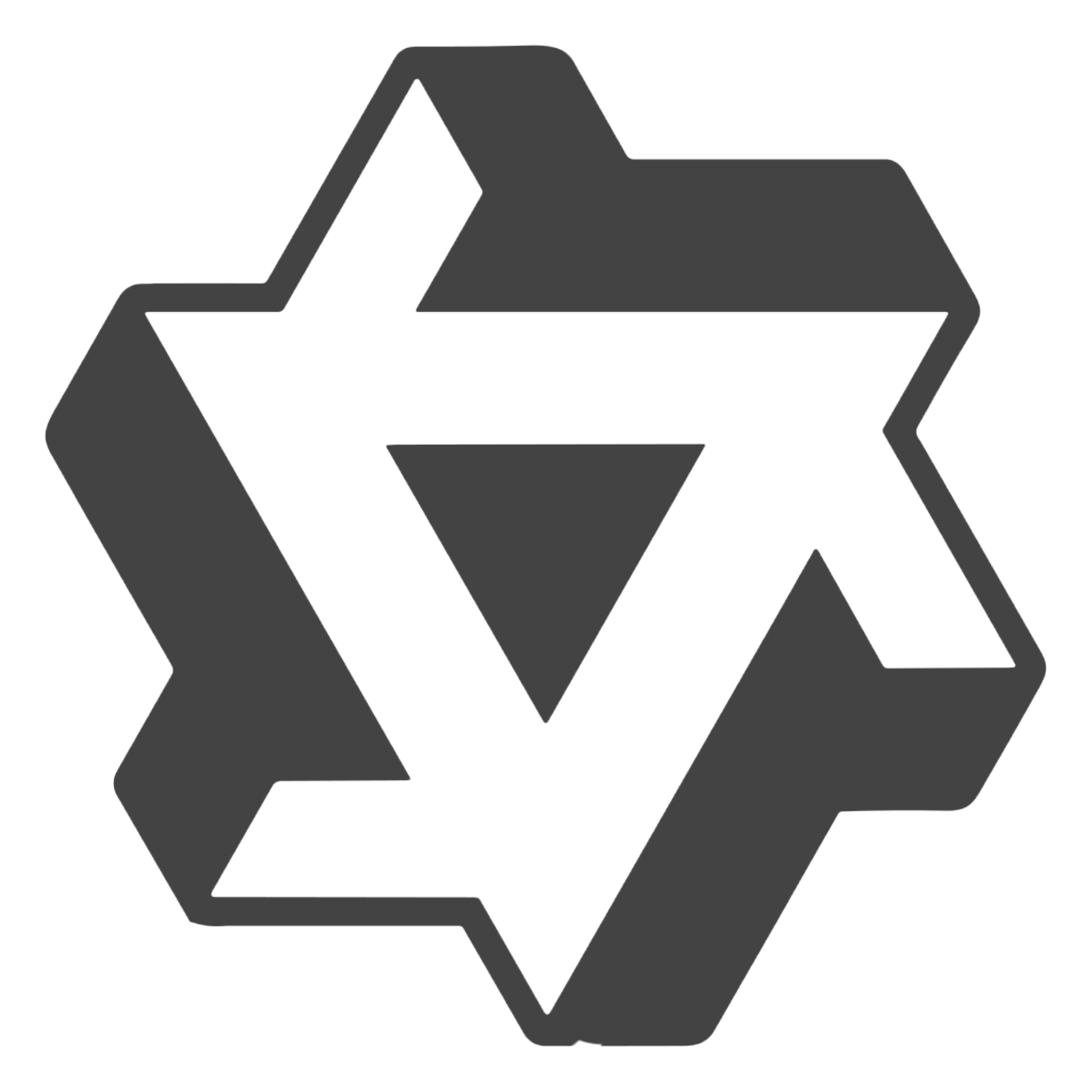}}\hspace{0.01em}}
\newcommand*\toy{\raisebox{-0.23em}{\includegraphics[width=1.2em]{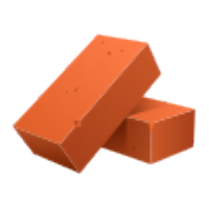}}\hspace{0.02em}}

\definecolor{dark_green_drawio}{HTML}{557543}
\definecolor{dark_red_drawio}{HTML}{990000}
\definecolor{mypurple}{HTML}{8B3BB5}
\definecolor{mygreen}{HTML}{769540}
\definecolor{myblue}{HTML}{5369C2}
\definecolor{mygrey}{HTML}{4A4F48}
\definecolor{inputseq}{HTML}{EED4D3}
\definecolor{parityseq}{HTML}{E8EDDF}
\definecolor{deepgreen}{HTML}{137E43}
\newcommand{\mystrut}{\rule[-0.3ex]{0pt}{1.65ex}} 

\makeatletter
\renewcommand{\maketitle}{\bgroup\setlength{\parindent}{0pt}
\begin{flushleft}
  \textbf{\@title}
  
  \@author
\end{flushleft}\egroup}
\makeatother

\title{Attention Heads of Large Language Models: A Survey}
\date{}

\author[1,4]{Zifan Zheng}
\author[1,4]{Yezhaohui Wang}
\author[2,4]{Yuxin Huang}
\author[1]{Shichao Song}
\author[3]{Mingchuan Yang}
\author[1]{Bo Tang}
\author[1]{Feiyu Xiong}
\author[1,*]{Zhiyu Li}

\affil[1]{Institute for Advanced Algorithms Research (IAAR), Shanghai, China}
\affil[2]{Institute for AI Industry Research (AIR), Tsinghua University, Beijing, China}
\affil[3]{Research Institute of China Telecom, Beijing, China}
\affil[4]{These authors contributed equally.}
\affil[*]{Correspondence: \href{mailto:lizy@iaar.ac.cn}{lizy@iaar.ac.cn}}

\usepackage[super,comma,sort&compress]{natbib}

\begin{document}

\maketitle

\section*{Summary}

Since the advent of ChatGPT, Large Language Models (LLMs) have excelled in various tasks but remain as black-box systems. \revise{Understanding the reasoning bottlenecks of LLMs has become a critical challenge, as these limitations are deeply tied to their internal architecture. Among these, attention heads have emerged as a focal point for investigating the underlying mechanics of LLMs.}
\revise{In this survey, we aim to demystify the internal reasoning processes of LLMs by systematically exploring the roles and mechanisms of attention heads. We first introduce a novel four-stage framework inspired by the human thought process:} Knowledge Recalling, In-Context Identification, Latent Reasoning, and Expression Preparation. Using this framework, we \revise{comprehensively} review existing research to identify and categorize the functions of specific attention heads. \revise{Additionally}, we \revise{analyze} the experimental methodologies used to discover these special heads, dividing them into two categories: Modeling-Free and Modeling-Required methods. \revise{We further summarize} relevant evaluation methods and benchmarks. Finally, we discuss the limitations of current research and propose several potential future directions.

\section*{Keywords}
Attention Head, Mechanistic Interpretability, Large Language Model (LLMs), Cognitive Neuroscience

\section{Introduction}
\begin{figure}[!ht]
    \centering
    \includegraphics[width=\linewidth]{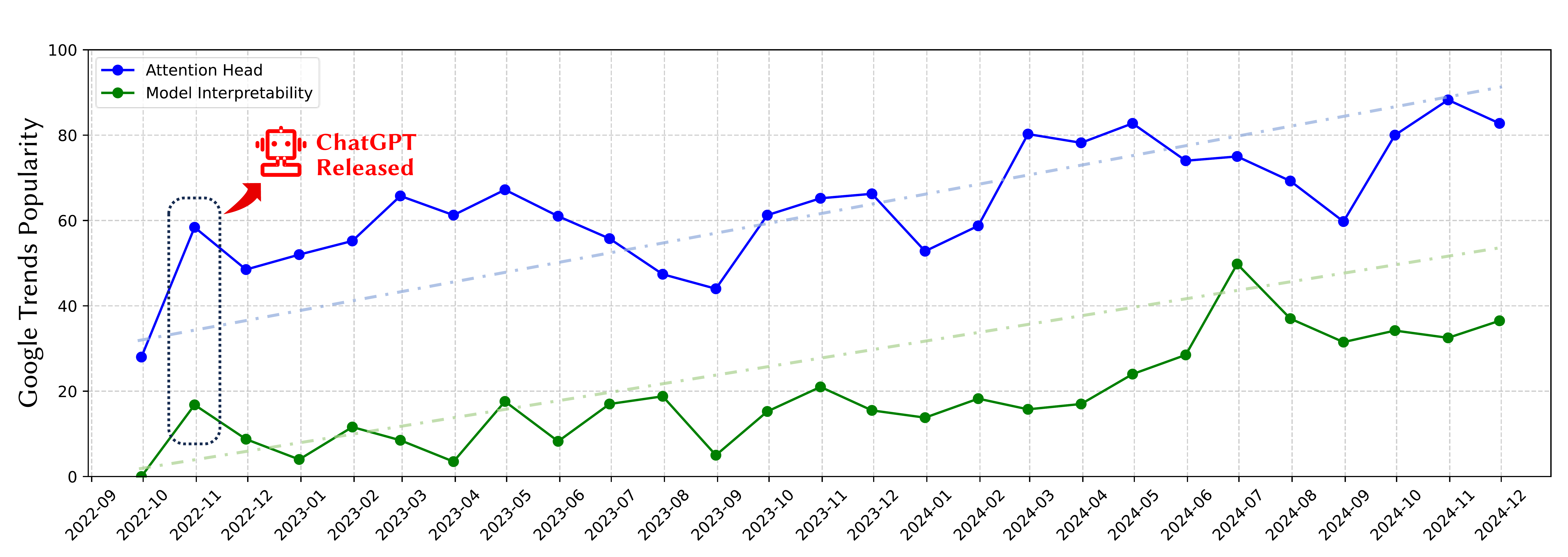}
    \caption{\revise{The global Google Trends Popularity of the keywords ``Attention Head'' and ``Model Interpretability''. The data retrieval date is December 4th, 2024.}}
    \label{fig:google_trends}
\end{figure}

The Transformer architecture \citep{AttentionIsAllYouNeed} has demonstrated outstanding performance across various tasks, such as Natural Language Inference and Natural Language Generation. However, it still retains the black-box nature inherent to Deep Neural Networks (DNNs).\citep{LLMblackbox_gilpin2018, LLMblackbox_lipton2018} As a result, many researchers have dedicated efforts to understanding the internal reasoning processes within these models, aiming to uncover the underlying mechanisms.\citep{DNN_Interp_Montavon} This line of research provides a theoretical foundation for models like BERT \citep{BERT_MODEL} and GPT \citep{GPT2_MODEL} to perform well in downstream applications. Additionally, in the current era where Large Language Models (LLMs) are widely applied, interpretability mechanisms can guide researchers in intervening in specific stages of LLM inference, thereby enhancing their problem-solving capabilities.\citep{AttnLookback_24_arXiv_MIT, MaskLayer_24_arXiv_NUDT, ji2024llm}
Among the components of LLMs, attention heads play a crucial role in the reasoning process. Particularly in recent years, attention heads within LLMs have garnered significant attention, as illustrated in Figure~\ref{fig:google_trends}. Numerous studies have explored attention heads with specific functions. This paper consolidates these research efforts, organizing and analyzing the potential mechanisms of different types of attention heads. Additionally, we summarize the methodologies employed in these investigations.

\begin{figure}[!ht]
    \centering
    \includegraphics[width=\linewidth]{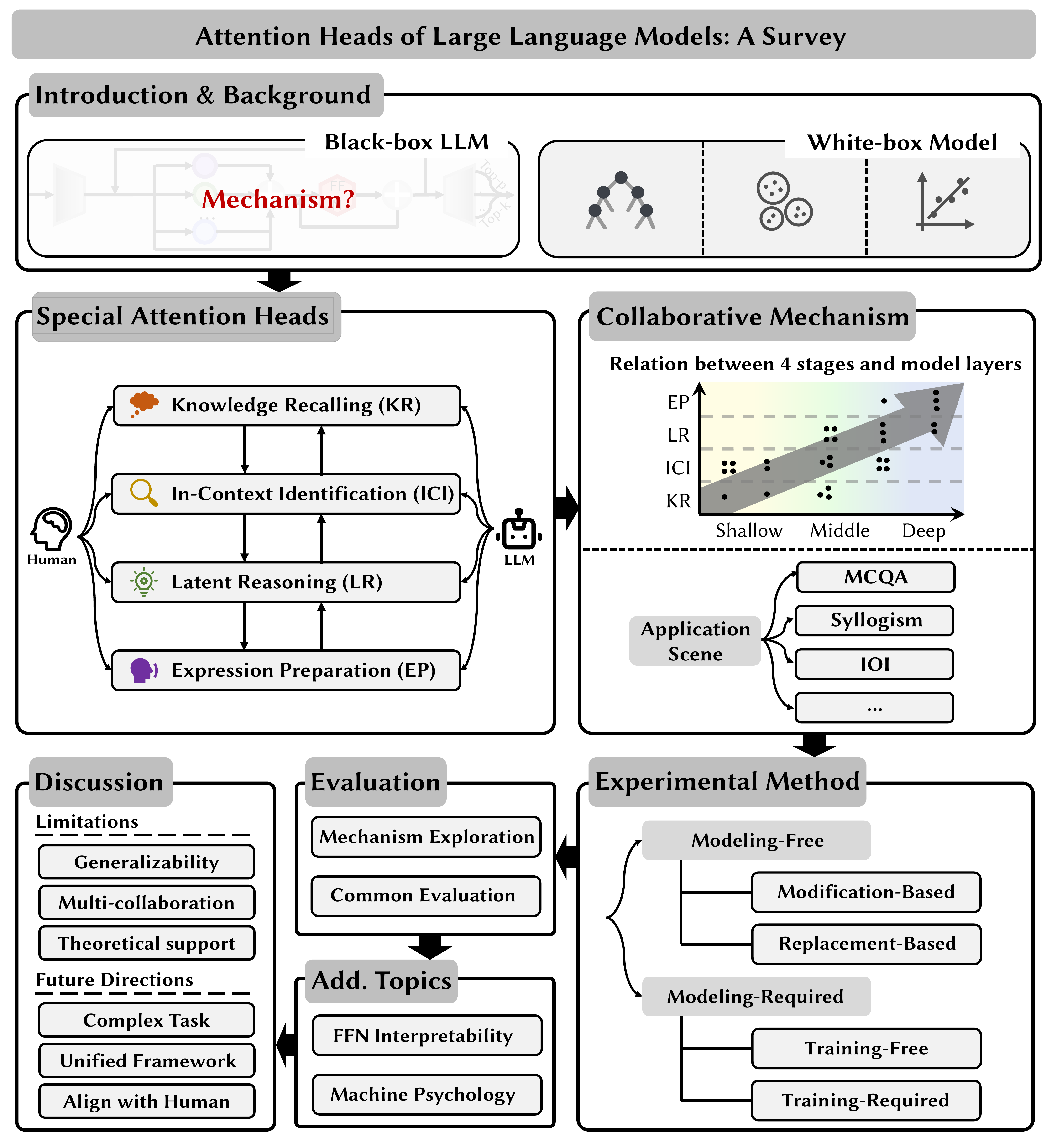}
    \caption{\revise{The framework of this survey.}}
    \label{fig:Overview}
\end{figure}

The logical structure and classification method of this paper are illustrated in Figure~\ref{fig:Overview}. We begin with the background of the problem in \nameref{sec:background}, where we present a simplified representation of the LLMs' structures (\nameref{subsec:Math}) and explain the related key terms (\nameref{subsec:KeyTerm}). In \nameref{sec:HeadOverview}, we first summarize the four stages of human thought processes from a cognitive neuroscience perspective and apply this framework to analyze the reasoning mechanisms of LLMs. Using this as our classification criterion, we categorize existing work on attention heads, identifying commonalities among heads that contribute to similar reasoning processes (from \nameref{subsec:KR} to \nameref{subsec:EP}) and exploring the collaborative mechanisms of heads functioning at different stages (\nameref{subsec:WorkTogether}).

Investigating the internal mechanisms of models often requires extensive experiments to validate hypotheses. To provide a comprehensive understanding of these methods, we summarize the current experimental methodologies used to explore special attention heads in \nameref{sec:DiscoveryExp}. We divide these methodologies into two main categories based on whether they require additional modeling: Modeling-Free (\nameref{subsec:ModelFree}) and Modeling-Required (\nameref{subsec:ModelRequired}).

In addition to the core sections shown in Figure~\ref{fig:Overview}, we summarize the evaluation tasks and benchmarks used in relevant studies in \nameref{sec:Evaluation}.
Furthermore, in \nameref{sec:other_tasks}, we compile research on the mechanisms of Feed-Forward Networks (FFNs) and Mechanical Interpretability to help deepen our understanding of LLM structures from multiple perspectives.
Finally, in \nameref{sec:Discussion}, we offer our insights on the current state of research in this field and outline several potential directions for future research.

In summary, the strengths of our work are:
\begin{itemize}
    \item \textbf{Focus on the latest research.} Although earlier researchers explored the mechanisms of attention heads in models like BERT, many of these conclusions are now outdated. This paper primarily focuses on highly popular LLMs, such as LLaMA and GPT, consolidating the latest research findings.
    \item \textbf{An innovative four-stage framework for LLM reasoning.} We have distilled key stages of human thought processes by integrating knowledge from cognitive neuroscience, psychology, and related fields. Furthermore, we have applied these stages as an analogy for LLM reasoning.
    \item \textbf{Detailed categorization of attention heads.} Based on the proposed four-stage framework, we classify different attention heads according to their functions within these stages, and we explain how heads operating at different stages collaborate to achieve alignment between humans and LLMs.
    \item \textbf{Clear summarization of experimental methods.} We provide a detailed categorization of the current methods used to explore attention head functions from the perspective of model dependency, laying a foundation for the improvement and innovation of experimental methods in future research.
\end{itemize}

\section{Out-of-scope topics} \label{sec:out_of_scope}
First, we need to clarify the boundaries of the topic reviewed in this paper. In other words, some works fall outside the scope of our focus.

\revise{As the latest research on attention head interpretability is primarily based on LLMs,} this paper focuses on the heads within current mainstream LLM architectures, specifically those with a \textbf{decoder-only structure}. As such, we do not discuss early studies related to the Transformer, such as those focusing on attention heads in BERT-based models.\citep{BertInterp_19_ACL,BertMarkov_19_ACL,BertAnalysis_21_AAAI_India}

Some studies on mechanistic interpretability propose holistic operational principles that encompass embeddings, attention heads, and MLPs. However, this paper focuses \textbf{exclusively on attention heads}. Consequently, we do not cover the roles of other components within the Transformer architecture from \nameref{sec:HeadOverview} to \nameref{sec:DiscoveryExp}; these are only briefly summarized in \nameref{sec:other_tasks}.

\section{Background} \label{sec:background}
\subsection{Mathematical representation of LLMs} \label{subsec:Math}
\revise{As mentioned in \nameref{sec:out_of_scope}, to facilitate the discussion in the subsequent sections, we first define the mathematical notations for the transformer layer of an LLM. Note that there are two main layer normalization methods in LLMs: Pre-Norm and Post-Norm.\citep{liu-etal-2020-understanding, xiong2020layer} However, since these are not the focus of this paper, we will \textbf{omit Layer Normalization} in this section.}

\begin{figure}[!ht]
    \centering
    \includegraphics[width=\linewidth]{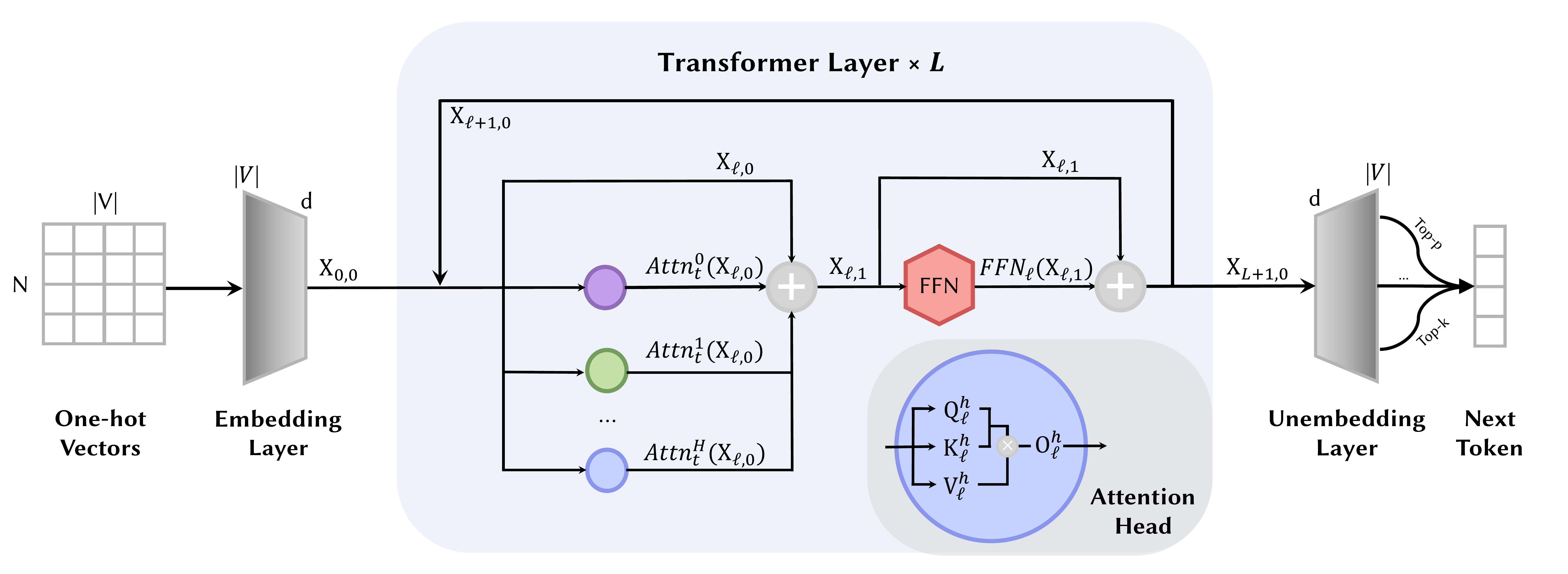}
    \caption{\revise{The overall structure of decoder-only LLMs.}}
    \label{fig:LLMStructure}
\end{figure}

As shown in Figure~\ref{fig:LLMStructure}, a model $\mathcal{M}$ consists of an embedding layer, $L$ identical \revise{transformer layers}, and an unembedding layer. The input to $\mathcal{M}$ are one-hot sentence tokens, with a shape of $\{0,1\}^{N \times |\mathcal{V}|}$, where $N$ is the length of the token sequence and $|\mathcal{V}|$ represents the vocabulary size.

After passing through the embedding layer, which applies semantic embedding $\mathbf{W_{E}} \in \mathbb{R}^{|\mathcal{V}| \times d}$ and positional encoding $\mathbf{P_{E}}$ (e.g., RoPE~\citep{ROPE_24_Neuro_Zhuiyi}), the one-hot matrix is transformed into the input $\mathbf{X}_{0,0} \in \mathbb{R}^{N \times d}$ for the first \revise{transformer layer}, where $d$ represents the dimension of the token embedding (or latent vector).

In the $\ell$-th $\ell \left(1\leq \ell \leq L\right)$ \revise{transformer layer}, there are two residual blocks.
\revise{The first residual block takes the input matrix $\mathbf{X}_{\ell,0} \in \mathbb{R}^{N \times d}$ and combines it with the output $\mathbf{X}_{\ell}^{\text{attn}}$—produced by a multi-head attention mechanism with $H$ attention heads—to compute $\mathbf{X}_{\ell,1}$ (as shown in Equation~\ref{equ:decoder0}).}
Subsequently, $\mathbf{X}_{\ell,1}$ serves as the input for the second residual block. Here, $Attn_{\ell}^{h}\left(\cdot\right)\  \left(1\leq \ell \leq L, 1\leq h \leq H\right)$ represents the computation function of the $h$-th attention head in the $\ell$-th layer (see Equation~\ref{equ:attention}), where $1 \leq h \leq H$.
\begin{equation} \label{equ:decoder0}
\begin{aligned}
    \mathbf{X}_{\ell}^{\text{attn}} &= \sum_{h=1}^{H}{Attn_{\ell}^{h}}\left(\mathbf{X}_{\ell,0}\right) \\
    \mathbf{X}_{\ell,1} &= \mathbf{X}_{\ell,0} + \mathbf{X}_{\ell}^{\text{attn}}
\end{aligned}
\end{equation}

Similarly, as shown in Equation~\ref{equ:decoder1}, the second residual block combines $\mathbf{X}_{\ell,1}$ with the output $\mathbf{X}_{\ell}^{\text{ffn}}$ obtained after passing through the FFN, yielding the final output $\mathbf{X}_{\ell+1,0}$ of the $\ell$-th decoder block. This output also serves as the input for the $\ell$+1-th decoder block. Here, $FFN_{\ell}\left(\cdot\right)$ consists of linear layers (and activation functions) such as GLU (Gated Linear Units), SwiGLU,\citep{GLU_2020_arXiv_Google} or MoEs.\citep{MOETransformer,SurveyMOE_24_arXiv_HKUST}
\begin{equation} \label{equ:decoder1}
\begin{aligned}
    \mathbf{X}_{\ell}^{\text{ffn}} &= FFN_{\ell}\left(\mathbf{X}_{\ell,1}\right) \\
    \mathbf{X}_{\ell+1,0} &= \mathbf{X}_{\ell,1} + \mathbf{X}_{\ell}^{\text{ffn}}
\end{aligned}
\end{equation}

Here, we will concentrate on the details of $Attn_{\ell}^{h}\left(\cdot\right)$. This function can be expressed using matrix operations.
Specifically, each layer's $Attn_{\ell}^{h}\left(\cdot\right)$ function corresponds to four low-rank matrices: $\mathbf{W_Q}_{\ell}^{h}, \mathbf{W_K}_{\ell}^{h}, \mathbf{W_V}_{\ell}^{h} \in \mathbb{R}^{d \times \frac{d}{H}}, \mathbf{O}_{\ell}^{h} \in \mathbb{R}^{\frac{d}{H} \times d}$. By multiplying $\mathbf{X}_{\ell,0}$ with $\mathbf{W_Q}_{\ell}^{h}$, the query matrix $\mathbf{Q}_{\ell}^{h} \in \mathbb{R}^{N \times \frac{d}{H}}$ is obtained. Similarly, the key matrix $\mathbf{K}_{\ell}^{h}$ and the value matrix $\mathbf{V}_{\ell}^{h}$ can be derived.
The function $Attn_{\ell}^{h}\left(\cdot\right)$ can then be expressed as Equation~\ref{equ:attention}.\citep{AttentionIsAllYouNeed}
\begin{equation} \label{equ:attention}
    Attn_{\ell}^{h}\left(\mathbf{X}_{\ell,0}\right) = \operatorname{softmax}\left(\mathbf{Q}_{\ell}^{h} \cdot \mathbf{K}_{\ell}^{h^\top}\right) \cdot \mathbf{V}_{\ell}^{h} \cdot \mathbf{O}_{\ell}^{h}
\end{equation}

\begin{figure}[htbp]
    \centering
    \includegraphics[width=\linewidth]{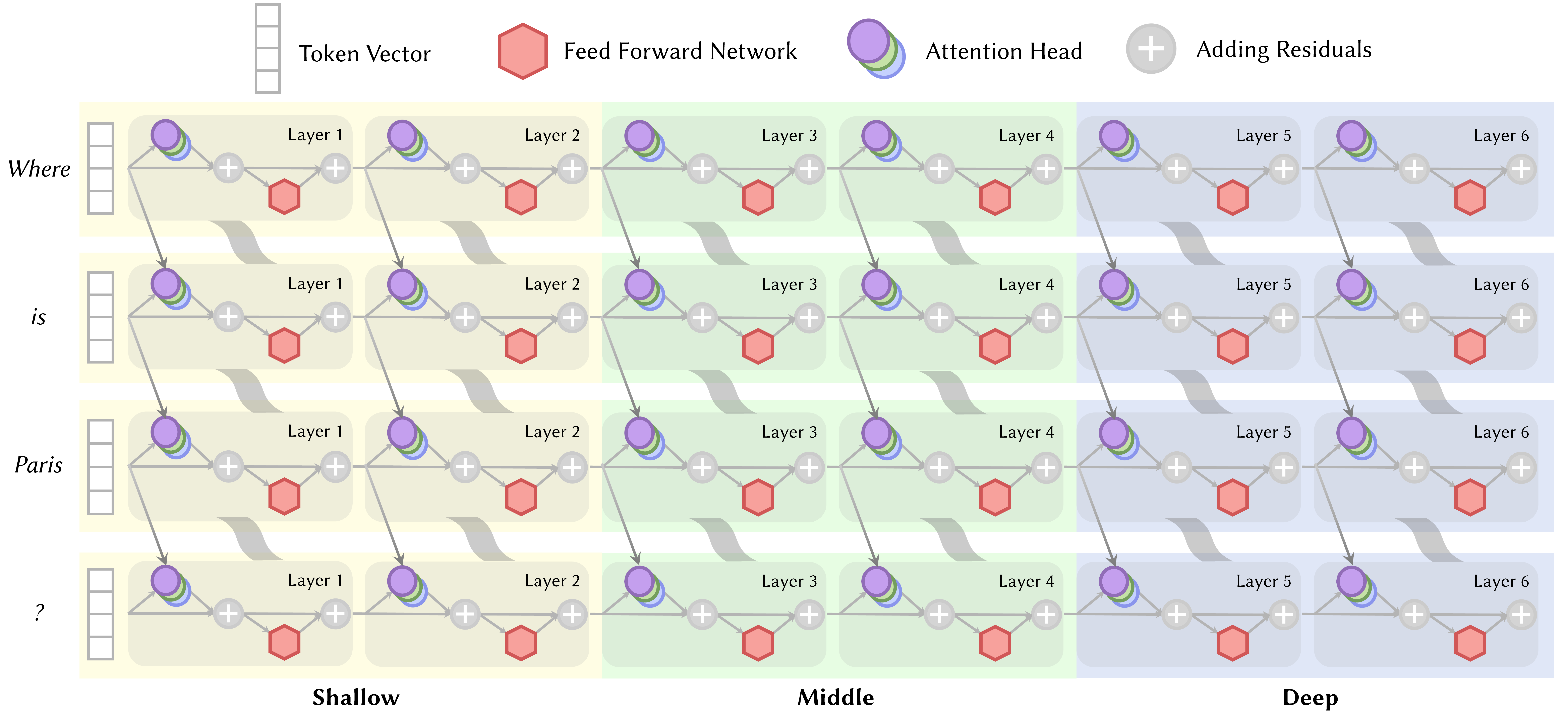}
    \caption{The diagram of residual streams. From the perspective of residual streams, the inference process of LLMs can be understood at a micro-level where attention heads access latent state matrices from several residual streams, as indicated by the gray arrows across layers in the diagram. At a macro-level, different residual streams control the information flow through attention heads, as shown by the gray wavy lines in the diagram.}
    \label{fig:ResidualStream}
\end{figure}

\subsection{Glossary of key terms} \label{subsec:KeyTerm}
\revise{This paper mentions several specialized terms that are fundamental to understanding and analyzing the reasoning mechanisms of LLMs. These terms are organized into two categories: conceptual frameworks, which provide theoretical abstractions for modeling LLM reasoning, and empirical analysis methods, which offer practical tools for experimentally probing and validating these frameworks. Below, we provide explanations for these key terms. For additional definitions related to model interpretability, please refer to the works of \citet{MIGlossary_22_blog_Neel}.}

\subsubsection{\revise{Conceptual frameworks}}
\paragraph{Circuits} Circuits are abstractions of the reasoning logic in deep models. The model $\mathcal{M}$ is viewed as a computational graph. There are two main approaches to modeling circuits. One approach treats the features in the latent space of $\mathcal{M}$ as nodes and the transitions between features as edges.\citep{OldCircuit_20_distill_OpenAI,CausalAbstract_21_NIPS_Stanford} The other approach views different components of $\mathcal{M}$, such as attention heads and neurons, as nodes; and the interactions between these components, such as residual connections, as edges.\citep{IOI_23_ICLR_Redwood} A circuit is a subgraph of $\mathcal{M}$. \revise{Researchers have discovered many important circuits, such as the Bias Circuit,\citep{GenderBias_20_NIPS_Salesforce} Knowledge Circuit,\citep{KnowledgeCircuit_24_arXiv_ZJU} and so on.}

\paragraph{Residual Stream} \revise{As shown in Figure~\ref{fig:ResidualStream}, each row in the figure can be viewed as a residual stream.} The residual stream after layer $\ell$ is the sum of the embedding and the outputs of all layers up to layer $\ell$, serving as the input to layer $\ell+1$. \citet{MathFrame_21_TCT_Anthropic} conceptualized the residual stream as a shared bandwidth through which information can flow. \revise{Different layers (or tokens) utilize this shared bandwidth, with lower layers (or previous tokens) writing information and higher layers (or subsequent tokens) reading it.}

\paragraph{\revise{QK Matrix \& OV Matrix}} \revise{We expand Equation~\ref{equ:attention} into Equation~\ref{equ:QKOV}. According to the study by \citet{MathFrame_21_TCT_Anthropic}, $\mathbf{W_Q}_{\ell}^h  \mathbf{W_K}_{\ell}^{h^\top}$ is referred to as the QK matrix (QK circuit), while $\mathbf{W_V}_{\ell}^h \mathbf{O}_{\ell}^h$ is referred to as the OV matrix (OV circuit). Specifically, the QK matrix enables the computation of attention scores between the $N$ tokens in $\mathbf{X}_{\ell, 0}$, thereby facilitating the reading of information from certain residual streams. Meanwhile, the OV matrix is responsible for writing the processed information back into the corresponding residual streams.}
\vspace{1.2em}
\revise{
\begin{equation}
\label{equ:QKOV}
\begin{aligned}
\operatorname{Attn}_{\ell}^h\left(\mathbf{X}_{\ell, 0}\right) 
&= \operatorname{softmax}\left(\left(\eqnmarkbox[mypurple]{QueryVector}{\mathbf{X}_{\ell, 0} \cdot \mathbf{W_Q}_{\ell}^h}\right) \cdot \left(\eqnmarkbox[mygreen]{KeyVector}{\mathbf{X}_{\ell, 0} \cdot \mathbf{W_K}_{\ell}^h}\right)\right)^\top \cdot \left(\eqnmarkbox[myblue]{ValueVector}{\mathbf{X}_{\ell, 0} \cdot \mathbf{W_V}_{\ell}^h}\right) \cdot \mathbf{O}_{\ell}^h\\
&= \operatorname{softmax}\left(\mathbf{X}_{\ell, 0} \cdot \eqnmarkbox[dark_red_drawio]{QKMatrix}{\mathbf{W_Q}_{\ell}^h  \mathbf{W_K}_{\ell}^{h^\top}} \cdot \mathbf{X}_{\ell, 0}^{\top} \right) \cdot \mathbf{X}_{\ell, 0} \cdot \eqnmarkbox[mygrey]{OVMatrix}{\mathbf{W_V}_{\ell}^h \mathbf{O}_{\ell}^h}
\annotate{above, left, label below}{QueryVector}{Query Vectors' Matrix}
\annotate{above, label below}{KeyVector}{Key Vectors' Matrix}
\annotate{above, label below}{ValueVector}{Value Vectors' Matrix}
\annotate[yshift=-0.5em]{below, label above}{QKMatrix}{QK Matrix}
\annotate[yshift=-0.5em]{below, label above}{OVMatrix}{OV Matrix}
\end{aligned}
\end{equation}
}

\subsubsection{\revise{Empirical analysis methods}}
\begin{figure}[htbp]
    \centering
    \includegraphics[width=0.85\linewidth]{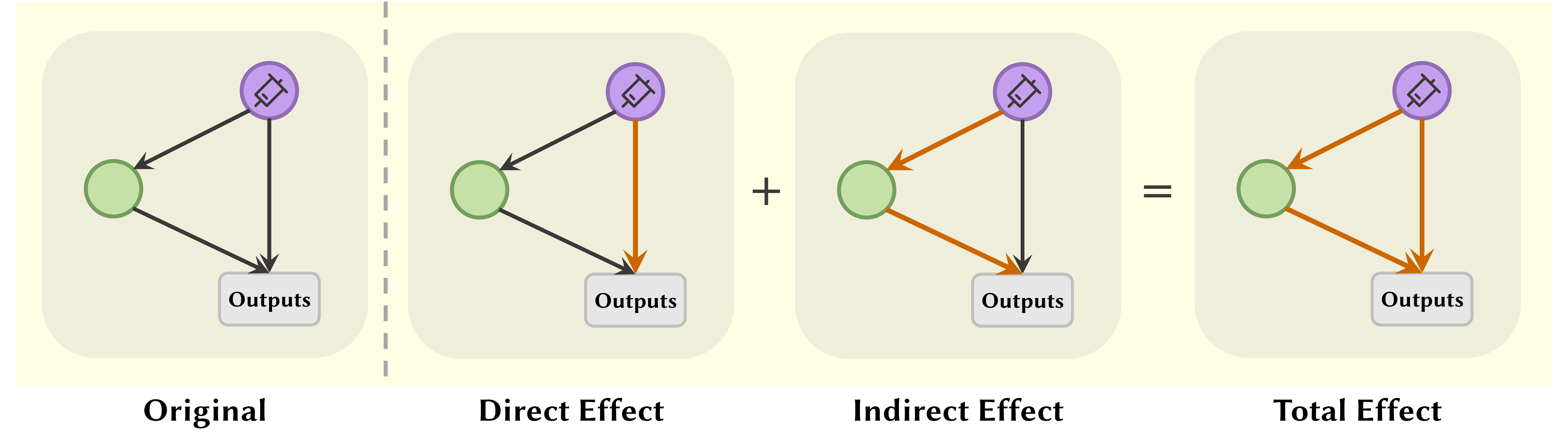}
    \caption{\revise{Three different types of calculating effects.}}
    \label{fig:ThreeEffect}
\end{figure}
\paragraph{\revise{Activation Patching}} \revise{Activation patching is aimed to analyze the impact of the modifications on the model's final decisions. It involves substituting activation values in specific layers of a model with alternatives—such as activations from different inputs, baseline values, or perturbed versions.} Specifically, three types of effects are considered: direct effect, indirect effect, and total effect, as illustrated in Figure~\ref{fig:ThreeEffect}.

\paragraph{\revise{Ablation study}} \revise{Ablation study and activation patching are conceptually related but differ in their methods of operation. Instead of replacing activations, it involves removing specific components of the LLM to observe how the output is affected.\citep{ActivationPatching_24_arXiv_Google} 
The key distinction between the two methods lies in their mechanism: activation patching modifies activations to simulate the logical replacement of a component, whereas ablation study physically removes the component entirely.}

\paragraph{Logit lens} When calculating effects like those shown in Figure~\ref{fig:ThreeEffect}, logit lens can quantify this effect. \revise{It is often used in conjunction with activation patching or ablation studies.} Specifically, it uses the unembedding layer to map an intermediate representation vector to the logits values of the vocabulary, allowing for the comparison of logits differences or other metrics. More details are in \href{https://colab.research.google.com/drive/1MjdfK2srcerLrAJDRaJQKO0sUiZ-hQtA}{the Colab notebook}.

\subsection{Existing related surveys}
To the best of our knowledge, there is \textbf{no survey} focused on the mechanisms of LLMs' attention heads. Specifically, \citet{SurveyDNNInner_23_SaTML_MIT} mainly discussed non-Transformer architectures, with little focus on attention heads. The surveys by \citet{SurveyMdedical_22_IEEE_Portugal, SurveyNeurAttn_21_arXiv_Brazil, SurveyAttentionModel_21_Linkedin, SurveydDLAttn_22_arixv_Netherland} cover older content, primarily focusing on the various attention computation methods that emerged during the early development of the Transformer. However, current LLMs still use the original scaled-dot product attention, indicating that many of the derived attention forms have become outdated. Although \citet{SurveyLLMInterp_24_arXiv} focused on the internal structure of LLMs, they only summarized experimental methodologies and overlooked research findings related to operational mechanisms.

\begin{table}[htbp]
\centering
\caption{\revise{Summary of the relationship between LLMs and human behaviors explored in existing studies.}}
\label{tab:viewpoints}
\resizebox{\textwidth}{!}{%
\revise{
\begin{tabular}{@{}ll@{}}
\toprule
\textbf{Research Paper}            & \multicolumn{1}{c}{\textbf{Viewpoints on the Relationship Between LLMs and Human (Brains)}}                                                                                                                                         \\ \midrule
\citet{ICSF_24_arXiv_IAAR}         & \begin{tabular}[c]{@{}l@{}}``Self-Feedback'' mechanism in LLMs mirrors \textbf{human metacognition} \citep{HumanMetacognition_18_Cambrige} by enabling models to evaluate and re-\\ fine their own reasoning.\end{tabular} \\
\citet{dasgupta2022language}       & The language model can exhibit many of the varied, \textbf{context-sensitive patterns} of human reasoning behavior.                                                                                                        \\
\citet{SAEHumanHeads_24_arXiv_MIT} & \begin{tabular}[c]{@{}l@{}}Different attention heads in LLMs exhibit specialized roles, analogous to the \textbf{modular organization} of human \\ brain regions.\end{tabular} 
\\
\citet{janik2023aspects}           & LLMs exhibit some human-like memory characteristics, such as \textbf{primacy} and \textbf{recency effects}. \\
\citet{schrimpf2021neural}         & \begin{tabular}[c]{@{}l@{}}Representations in Transformers show significant similarity to human brain neural activities during language\\ tasks, particularly in terms of \textbf{predictive processing} (Errors flows bottom-up to adjust the model).\end{tabular}     \\
\citet{marjieh2024large}           & \begin{tabular}[c]{@{}l@{}}The attention distributions of LLMs for implicit semantic relations in language closely align with human res-\\ ponse patterns in \textbf{perceptual tasks}.\end{tabular}                       \\
\citet{mischler2024contextual}     & The attention mechanism may partially reflect the brain's \textbf{predictive coding theory} \citep{PredictiveCoding_21}.                                                                                                                                             \\ \bottomrule
\end{tabular}%
}
}
\end{table}
\section{Overview of special attention heads} \label{sec:HeadOverview}
Previous research has shown that the decoder-only architecture described in \nameref{sec:background} follows the Scaling Law, and it exhibits emergent abilities once the number of parameters reaches a certain threshold.\citep{ScalingLaw_20_arXiv_OpenAI, ScalingLaw_21_arXiv_Stanford} Many LLMs that have emerged subsequently demonstrate outstanding performance in numerous tasks, even close to humans. However, researchers still do not fully understand why these models are able to achieve such remarkable results. To address this question, recent studies have begun to delve into the internal mechanisms of LLMs, focusing on their fundamental structure—a neural network composed of multi-attention heads and FFNs.

We have observed that many studies concentrate on the functions of attention heads, attempting to explain their reasoning processes. Additionally, several researchers have drawn parallels of reasoning methods between LLMs and human, \revise{as illustrated in Table~\ref{tab:viewpoints}. These findings suggest that certain research insights from studies of the human brain may be transferable to the study of attention heads. Therefore, in this section, we first summarize a four-stage framework inspired by human cognitive paradigms and use it as a guiding method to classify the functions of different attention heads.}

\subsection{\revise{How does the human brain \& attention head think?}}
\revise{As shown in Table~\ref{tab:viewpoints}, the role of an attention head, as its name suggests, is quite analogous to the functions of the human brain. In some representative earlier works, the OAR model abstracts human brain knowledge and information into a graph composed of objects, attributes, and relations.\citep{OAR_model} Based on this abstraction, \citet{Cognitive_OAR} proposed a mathematical model of problem solving. Specifically, the solver's brain first utilizes its own OAR model to identify the content of the problem, distinguishing the objects and attributes within it, and constructs a sub-OAR model accordingly. Then, the solver combines their knowledge to search for potential solution goals and solution paths, evaluating these candidate solutions. If the evaluation results are unsatisfactory, the solver iteratively explores and evaluates solutions until suitable ones are found. Ultimately, the result of problem solving is represented as a part of the relations in the sub-OAR model.}

\revise{Similarly, the ACT-R model, which consists of five modules—Perception (P), Working Memory (WM), Procedural Memory (PM), Declarative Memory (DM), and Motor (M)—highlights the interaction between various modules in human cognition.\citep{ACTR_anderson} The P module receives environmental inputs (e.g., visual or auditory information) and transmits them to the WM module. WM retrieves condition-action rules stored in PM in an if-then format to generate the next action. If additional knowledge is required, WM retrieves it from DM. Finally, the action is executed through the M module.\citep{whitehill2013understanding,laird2022analysis}}

\begin{figure}[htbp]
    \centering
    \includegraphics[width=0.8\linewidth]{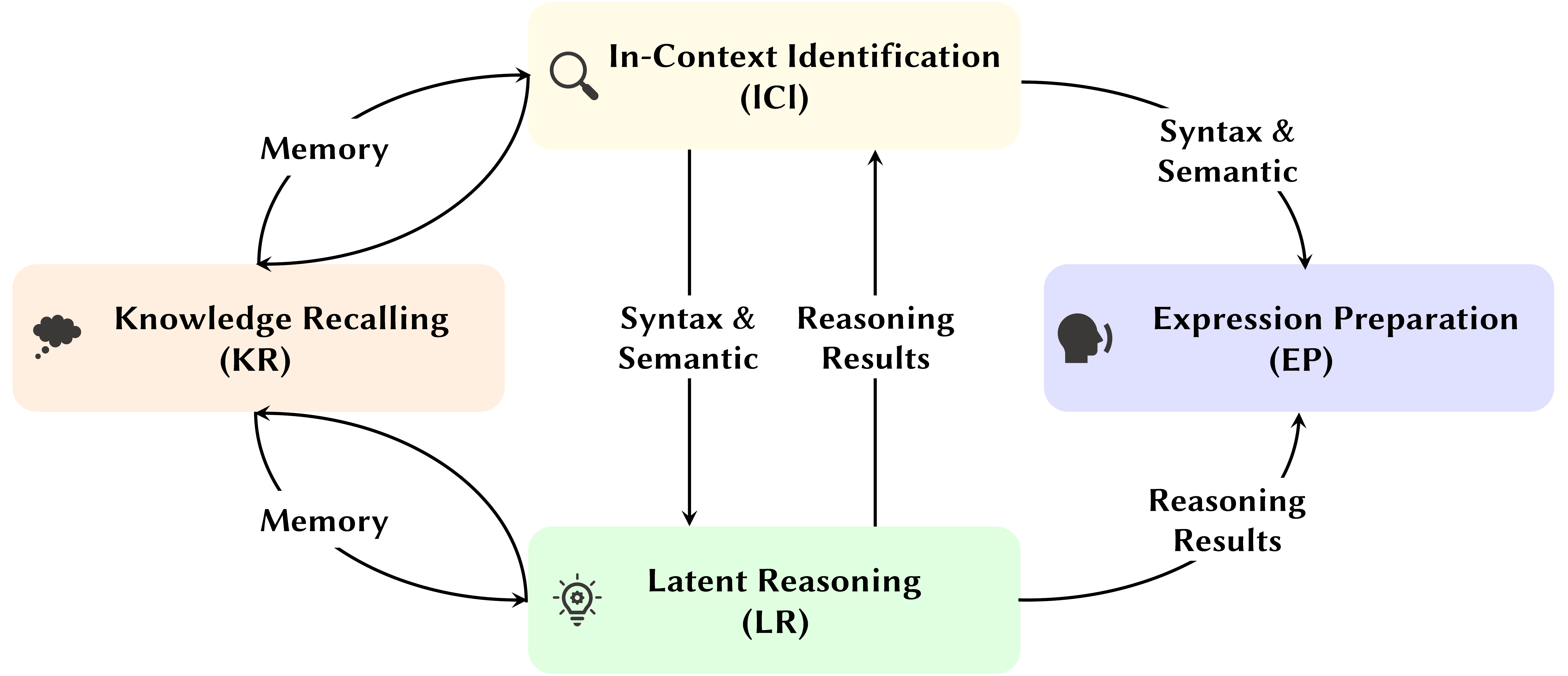}
    \caption{\revise{The four-stage framework of human thinking and LLM reasoning. The relationship between these four stages is not a linear progression but rather a graph-like transformation. Both humans and LLMs iteratively retrieve internal knowledge, observe the problem, and reason to arrive at the final answer.}}
    \label{fig:four_steps}
\end{figure}

\revise{In summary, these studies center on how humans retrieve knowledge, perceive and understand problems or environments, and conceive and execute actions. Inspired by these works, we propose a more universally applicable four-stage framework for describing the process by which the human brain solves specific problems:} Knowledge Recalling (KR), In-Context Identification (ICI), Latent Reasoning (LR), and Expression Preparation (EP). These four stages can interact with and transition between one another, as illustrated in Figure~\ref{fig:four_steps}.

When solving a problem, humans first need to recall the knowledge they have learned that is relevant to the issue at hand. This process is known as \textbf{Knowledge Recalling (KR)}. During this stage, the hippocampus integrates memories into the brain's network \citep{SquireMemory} and activates different types of memories as needed \revise{with the help of dynamic associations.\citep{MemoryRecall,sartori2023language}} 
Confronted with the specific text of the problem, humans need to perform \textbf{In-Context Identification (ICI)}. This means that the brain not only focuses on the overall structural content of the text \citep{GeneralStructure} but also parses the syntactic \citep{Syntax} and semantic \citep{Semantic} information embedded within it.

Once the brain has acquired the aforementioned textual and memory information, it attempts to integrate this information to derive conclusions, a process known as \textbf{Latent Reasoning (LR)}. This stage primarily includes arithmetic operations \citep{ReasoningNumber} and logical inference \citep{ReasoningLogic}.
Finally, the brain needs to translate the reasoning results into natural language, forming an answer that can be expressed verbally. This is the \textbf{Expression Preparation (EP)} stage. At this point, the brain bridges the gap between ``knowing'' and ``saying''.\citep{LanguageExpress}

As indicated by the arrows in Figure~\ref{fig:four_steps}, these four stages are not executed in a strictly one-direction fashion when humans solve problems; rather, they can jump and switch between each other. For example, the brain may ``cycle'' through the identification of contextual content (the ICI stage) and then retrieve relevant knowledge based on the current context (the KR stage). Similarly, if latent reasoning cannot proceed further due to missing information, the brain may return to the Knowledge Recalling and In-Context Identification stages to gather more information.

We will now draw an analogy between these four steps and the mechanisms of attention heads, as depicted in Figure~\ref{fig:head_taxnomomy}. Previous research has shown that LLMs possess strong contextual learning abilities and have many practical applications.\citep{FewShotLearners} As a result, much of the work on interpretability has focused on the ability of LLMs to capture and reason about contextual information. Consequently, the functions of currently known special attention heads are primarily concentrated in the ICI and LR stages, while fewer attention heads operate in the KR and EP stages.

\definecolor{hidden-draw}{RGB}{128, 128, 128}

\tikzstyle{my-box}=[
    rectangle,
    draw=hidden-draw,
    rounded corners,
    align=left,
    text opacity=1,
    minimum height=2.5em,
    minimum width=5em,
    inner sep=2pt,
    fill opacity=.8,
    line width=0.8pt,
]

\tikzstyle{leaf-head}=[my-box, minimum height=2.5em,
    draw=gray!80, 
    fill=gray!35,  
    text=black, font=\normalsize,
    inner xsep=2pt,
    inner ysep=4pt,
    line width=0.8pt,
]

\tikzstyle{leaf-task}=[my-box, minimum height=2.5em,
    draw=red!80, 
    fill=red!20,  
    text=black, font=\normalsize,
    inner xsep=2pt,
    inner ysep=4pt,
    line width=0.8pt,
]

\tikzstyle{leaf-paradigms}=[my-box, minimum height=2.5em,
    draw=orange!70, 
    fill=orange!15,  
    text=black, font=\normalsize,
    inner xsep=2pt,
    inner ysep=4pt,
    line width=0.8pt,
]
\tikzstyle{leaf-others}=[my-box, minimum height=2.5em,
    draw=yellow!80, 
    fill=yellow!15,  
    text=black, font=\normalsize,
    inner xsep=2pt,
    inner ysep=4pt,
    line width=0.8pt,
]
\tikzstyle{leaf-other}=[my-box, minimum height=2.5em,
    draw=green!80, 
    fill=green!15,  
    text=black, font=\normalsize,
    inner xsep=2pt,
    inner ysep=4pt,
    line width=0.8pt,
]
\tikzstyle{leaf-application}=[my-box, minimum height=2.5em,
    draw=blue!80, 
    fill=blue!15,  
    text=black, font=\normalsize,
    inner xsep=2pt,
    inner ysep=4pt,
    line width=0.8pt,
]

\tikzstyle{modelnode-task}=[my-box, minimum height=2.5em,
    draw=red!80, 
    fill=red!20,  
    text=black, font=\normalsize,
    inner xsep=2pt,
    inner ysep=4pt,
    line width=0.8pt,
]
\tikzstyle{modelnode-paradigms}=[my-box, minimum height=2.5em,
    draw=orange!70, 
    fill=orange!15,  
    text=black, font=\normalsize,
    inner xsep=4pt,
    inner ysep=8pt,
    line width=1.5pt,
]
\tikzstyle{modelnode-others}=[my-box, minimum height=2.5em,
    draw=yellow!80, 
    fill=yellow!15,  
    text=black, font=\normalsize,
    inner xsep=4pt,
    inner ysep=8pt,
    line width=1.5pt,
]
\tikzstyle{modelnode-other}=[my-box, minimum height=2.5em,
    draw=green!80, 
    fill=green!15,  
    text=black, font=\normalsize,
    inner xsep=4pt,
    inner ysep=8pt,
    line width=1.5pt,
]
\tikzstyle{modelnode-application}=[my-box, minimum height=2.5em,
    draw=blue!80, 
    fill=blue!15,  
    text=black, font=\normalsize,
    inner xsep=4pt,
    inner ysep=8pt,
    line width=1.5pt,
]

\begin{figure*}[!ht]
    \centering
    \resizebox{1\textwidth}{!}
    {
        \begin{forest}
            for tree={
                grow=east,
                reversed=true,
                anchor=base west,
                parent anchor=east,
                child anchor=west,
                base=left,
                font=\normalsize,
                rectangle,
                draw=hidden-draw,
                rounded corners,
                align=left,
                minimum width=1em,
                edge+={darkgray, line width=1pt},
                s sep=10pt,
                inner xsep=0pt,
                inner ysep=3pt,
                line width=0.8pt,
                ver/.style={rotate=90, child anchor=north, parent anchor=south, anchor=center},
            }, 
            [
                \textbf{Special Attention Heads},leaf-head, ver   
                [
                    \textbf{ \S \nameref{subsec:KR}}, leaf-paradigms, text width=15em
                    [
                        \ \textbf{General Tasks}, leaf-paradigms, text width=11em
                        [\textbf{ \toy{}Associative Memories}~\citep{AssociativeMemory_23_NIPS_Meta,MemoryMath_24_arXiv_MILES}{, }\textbf{\llama{}\pythia{}\gpt{}Memory Head}~\citep{KnowledgeConflict_24_arXiv_UCAS},  modelnode-paradigms, text width=39.5em]
                    ]
                    [
                       \ \textbf{Specific Tasks}, leaf-paradigms, text width=11em
                         [\textbf{ \gemma{}Constant Head}{ (MCQA)}~\citep{CorrectLetterHead_23_arXiv_DeepMind}{, }\textbf{\gemma{}Single Letter Head}{ (MCQA)}~\citep{CorrectLetterHead_23_arXiv_DeepMind}{, }\textbf{\gemma{}\mistral{}\llama{}\qwen{}\gpt{}Negative }\\\textbf{ Head (BDT)}~\citep{NegativeHead_24_arXiv_SNU},                          modelnode-paradigms, text width=39.5em]
                    ]
                ]
                [
                    \textbf{ \S \nameref{subsec:ICI}}, leaf-others,text width=15em
                    [
                        \ \textbf{Overall Structure}, leaf-others, text width=11.8em
                        [\textbf{ \pythia{}\gpt{}Previous Head}~\citep{InductionHeads_22_TCT_Anthropic,PreviousHead_23_AIForum_Google}{, }\textbf{\gpt{}Positional Head}~\citep{InformationFlow_24_arXiv_Meta,SpecialHead_19_ACL_Russia,PositionalHead_18_ACL_Helsinki}{, }\textbf{\toy{}Rare Words Head}~\citep{SpecialHead_19_ACL_Russia}{, } \textbf{\gpt{}Dup-}\\\textbf{ licate Head}~\citep{IOI_23_ICLR_Redwood}{, }\textbf{\yi{}\mistral{}\llama{}\qwen{}Retrieval head}~\citep{RetrievalHead_24_arXiv_PKU}{, }\textbf{\llama{}Global Retrieval head}~\citep{RetrievalHead_24_arXiv_Huawei}, modelnode-others, text width=38.7em]
                    ]
                    [
                         \ \textbf{Syntactic Information}, leaf-others, text width=11.8em
                         [\textbf{ \gpt{}Subword Merge Head}~\citep{InformationFlow_24_arXiv_Meta,SubwordHead_19_ACL_Portugal}{, }\textbf{\toy{}Syntactic Head}~\citep{SpecialHead_19_ACL_Russia,SyntacticHead_23_arXiv_NYU}{, }\textbf{\gpt{}Negative Name Mover}\\\textbf{ Head}~\citep{IOI_23_ICLR_Redwood}{, }\textbf{\llama{}\gpt{}Mover Head}~\citep{KnowledgeCircuit_24_arXiv_ZJU}{, }\textbf{\gpt{}Name Mover Head}~\citep{CopySupression_23_arXiv_Google}{, }\textbf{\gpt{}Backup Name Mover}\\\textbf{ Head}~\citep{CopySupression_23_arXiv_Google}{, }\textbf{\gpt{}Letter Mover Head}~\citep{AcronymPredict_24_arXiv_Alicante}, modelnode-others, text width=38.7em]
                    ]
                    [
                        \ \textbf{Semantic Information}, leaf-others, text width=11.8em
                        [\textbf{ \gpt{}Context Head}~\citep{KnowledgeConflict_24_arXiv_UCAS}{, }\textbf{\gemma{}Content Gatherer Head}~\citep{CorrectLetterHead_23_arXiv_DeepMind,ColorObject_24_ICLR_BrownU}{, }\textbf{\gpt{}Sentiment Summarizer}~\citep{Sentiment_23_arXiv_EleutherAI}{, }\\\textbf{ \internlm{}Semantic Induction Head}~\citep{Semantic_24_arXiv_SJTU}{, }\textbf{\pythia{}Subject Head}{ \& }\textbf{\llama{}\pythia{}\gpt{}Relation Head}~\citep{FactualRecall_24_arXiv_Independent,KnowledgeCircuit_24_arXiv_ZJU}, modelnode-others,text width=38.7em]
                    ]
                ]
                [
                    \textbf{ \S \nameref{subsec:LR}}, leaf-other,text width=15em
                    [
                        \ \textbf{In-context Learning}, leaf-other, text width=10.5em
                        [
                            \textbf{ \gpt{}Summary Reader}~\citep{Sentiment_23_arXiv_EleutherAI}{, }\textbf{\llama{}\gpt{}Function Vector}~\citep{FunctionVector_24_ICLR_NEU}{, }\textbf{\yi{}\mistral{}\llama{}\qwen{}\gpt{}Induction Head}~\citep{InductionHeads_22_TCT_Anthropic,Markov_24_arXiv_Harvard,FSL_24_ICML_UCL,HumanMemory_24_arXiv_UCSD,InductionHead_24_arXiv_UoA,InductionHead_24_ICLR_Princeton,InductionHead_24_ICML_MIT}, modelnode-other, text width=40em
                        ]
                    ]
                    [
                        \ \textbf{Effective Reasoning}, leaf-other, text width=10.5em
                        [
                            \textbf{ \llama{}Truthfulness Head}~\citep{ITI_23_NIPS_harvard,NL-ITI_24_arXiv-Samsung}{, }\textbf{\gemma{}\llama{}Accuracy Head}~\citep{CrossLingual_24_SIGIR_UCAS,LoFiT_24_arXiv_UT}{, }\textbf{\llama{}Consistency Head}~\citep{SemanticConsistency_24_ACL_TJU}{, }\textbf{\gpt{}Vul-}\\\textbf{ nerable Head}~\citep{VulnerableHead_24_arXiv_Alicante}, modelnode-other, text width=40em
                        ]
                    ]
                    [
                        \ \textbf{Task-Specific Reas.}, leaf-other, text width=10.5em
                        [
                            \textbf{ \gemma{}Correct Letter Head}{ (MCQA)}~\citep{CorrectLetterHead_23_arXiv_DeepMind}{, }\textbf{\toy{}Iteration Head}{ (Sequence)}~\citep{IterationHead_24_arXiv_Meta}{, }\textbf{\pythia{}\gpt{}\llama{}Successor}\\\textbf{ Head}{ (Ordinal)}~\citep{SuccessorHead_24_ICLR_Cambridge}{, }\textbf{\gpt{}Inhibition Head}{ (Term)}~\citep{IOI_23_ICLR_Redwood,HeadCooperation_24_arXiv_UoM}, modelnode-other, text width=40em
                        ]
                    ]
                ]
                [
                    \textbf{ \S \nameref{subsec:EP}}, leaf-application,text width=15em
                    [
                        \ \textbf{Information Aggregation}, leaf-application, text width=12.5em
                        [
                            \textbf{ \pythia{}Mixed Head}~\citep{FactualRecall_24_arXiv_Independent}, modelnode-application, text width=38em
                        ]
                    ]
                    [
                        \ \textbf{Signal Amplification}, leaf-application, text width=12.5em
                        [
                            \textbf{ \gemma{}Amplification Head}~\citep{CorrectLetterHead_23_arXiv_DeepMind}{, }\textbf{\llama{}Correct Head}~\citep{CorrectHead_24_arXiv_Allen}, modelnode-application, text width=38em
                        ]
                    ]
                    [
                        \ \textbf{Instruction Alignment}, leaf-application, text width=12.5em
                        [
                            \textbf{ \llama{}Coherence Head}~\citep{CrossLingual_24_SIGIR_UCAS}{, }\textbf{\llama{}\gpt{}Faithfulness Head}~\citep{FaithfulCoT_24_ICML_Harvard}, modelnode-application, text width=38em
                        ]
                    ]
                ]
            ]
        \end{forest}
    }
    \caption{\revise{Taxonomy of special attention heads in language models. The icons before each head indicate the specific LLM architectures where the head was discovered. \internlm{}: InternLM series. \yi{}: Yi series. \gemma{}: Gemma series. \mistral{}: Mistral series. \llama{}: Llama series. \qwen{}: Qwen series. \pythia{}: Pythia series. \gpt{}: GPT series. \toy{}: Toy models, such as two-layer decoder-only transformers.}}
    \label{fig:head_taxnomomy}
\vspace{-0.3cm}
\end{figure*}
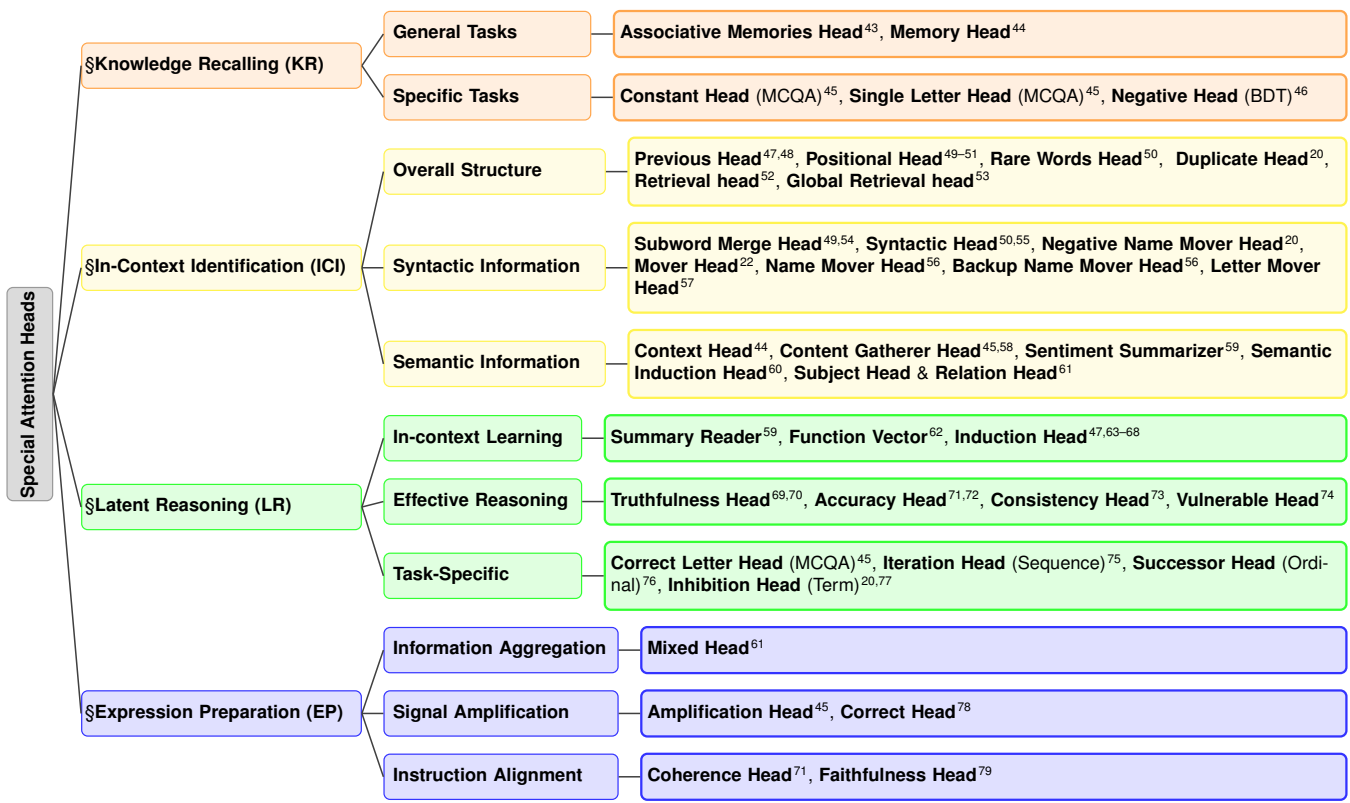

\subsection{Knowledge Recalling (KR)} \label{subsec:KR}
For LLMs, most knowledge is learned during the training or fine-tuning phases, which is embedded in the model's parameters. This form of knowledge is often referred to as LLMs' ``parametric knowledge''. Similar to humans, certain attention heads in LLMs recall this internally stored knowledge—such as common sense or domain-specific expertise—to be used in subsequent reasoning. \revise{These heads typically retrieve knowledge by making initial guesses or associating based on specific content within the context, injecting the memory information into the residual stream as initial data or supplementary information. A brief summary of their functionalities is shown in Table \ref{tab:KR_Head}.}

\revise{In \textbf{general tasks}, \citet{AssociativeMemory_23_NIPS_Meta} identified that certain attention heads can give rise to \textit{associative memories}, progressively storing and retrieving knowledge during the model's training phase. The weight matrices of these heads can be viewed as a weighted sum of the outer products of various vectors (e.g., input-output vectors or key-value vectors). Through their processing, these heads filter out noise from a superposed activation state while preserving essential features. Furthermore, as the embedding dimension $d$ increases, they become more adept at refining relevant information and linking it to useful memories.\citep{MemoryMath_24_arXiv_MILES}
The so-called \textit{Memory Head} is capable of retrieving content related to the current problem from the model’s parametric knowledge.\citep{KnowledgeConflict_24_arXiv_UCAS} This retrieved content could be knowledge learned during pre-training or experience accumulated during previous reasoning processes. Specifically, shallow FFNs enrich the semantics of entities present in the problems. Based on this enriched information, the Memory Head recalls attributes associated with these entities and writes them back into the residual stream.}

In \textbf{specific task scenarios}, such as when LLMs tackle Multiple Choice Question Answering (MCQA) problems, \revise{the answer is typically an option letter (e.g., A/B/C/D) rather than a short text}. In these cases, they may initially use \textit{Constant Head} to evenly distribute attention scores across all options. Alternatively, they might use \textit{Single Letter Head} to assign a higher attention score to one option while giving lower scores to others, thereby capturing all potential answers.\citep{CorrectLetterHead_23_arXiv_DeepMind}
In addition, in the context of Binary Decision Tasks (BDT), \revise{which are problems where the solution space is discrete and contains only two options, such as yes-no questions or answer verification,} \citet{NegativeHead_24_arXiv_SNU} found that LLMs often exhibit a negative bias when handling such tasks. This could be because the model has learned a significant amount of negative expressions related to similar tasks from prior knowledge during training. Consequently, when the model identifies a given text as a binary task, a \textit{Negative Head} may ``preemptively'' choose the negative answer due to this prior bias.
\begin{table}[htbp]
\centering
\caption{\revise{Key Attention Heads in Knowledge Recalling (KR).}}
\label{tab:KR_Head}
\resizebox{\textwidth}{!}{%
\revise{
\begin{tabular}{@{}lllc@{}}
\toprule
\multicolumn{1}{c}{\textbf{Head Name}} & \multicolumn{1}{c}{\textbf{Input Feature}} & \multicolumn{1}{c}{\textbf{Output Feature}}       & \textbf{Layer Distribution} \\ \midrule
Memory Head                            & User context \& Intermediate results     & Relevant parametric knowledge injected            & Shallow / Middle                      \\
Constant Head                          & All options in multiple-choice tasks       & Uniformly distributed attention scores            & Middle                     \\
Single Letter Head                     & Answer options   & Focused attention on a single candidate           & Middle                     \\
Negative Head                          & Binary decision task context               & Bias attention scores toward negative expressions & Middle                      \\ \bottomrule
\end{tabular}%
}
}
\end{table}

\subsection{In-Context Identification (ICI)} \label{subsec:ICI}
Understanding the in-context nature of a problem is one of the most critical processes to effectively address it. Just as humans read a problem statement and quickly pick up on various key pieces of information, some attention heads in LLMs also focus on these elements. Specifically, attention heads that operate during the ICI stage use their QK matrices to focus on and identify overall structural, syntactic, and semantic information within the in-context. This information is then written into the residual stream via OV matrices.

\subsubsection{Overall Structural Information Identification} \label{sucsubsec:overallstructure}
Identifying the overall structural information within a context mainly involves LLMs attending to content in special positions or with unique occurrences in the text.
\textit{Previous Head} \citep{InductionHeads_22_TCT_Anthropic,PreviousHead_23_AIForum_Google} and \textit{Positional Head}~\citep{InformationFlow_24_arXiv_Meta,SpecialHead_19_ACL_Russia,PositionalHead_18_ACL_Helsinki} attend to the positional relationships within the token sequence. They capture the embedding information of the current token and the previous token.
\textit{Rare Words Head} focuses on tokens that appear with the lowest frequency, emphasizing rare or unique tokens.\citep{SpecialHead_19_ACL_Russia}
\textit{Duplicate Head} excels at capturing repeated content within the context, giving more attention to tokens that appear multiple times.\citep{IOI_23_ICLR_Redwood}

Besides, as LLMs can gradually handle long texts, this is also related to the ``Needle-in-a-Haystack'' capability of attention heads. \textit{(Global) Retrieval Head} can accurately locate specific tokens in long texts.\citep{RetrievalHead_24_arXiv_PKU, GlobalAttention_24_arXiv_THU, RetrievalHead_24_arXiv_Huawei} These heads enable LLMs to achieve excellent reading and in-context retrieval capabilities.

\subsubsection{Syntactic Information Identification} \label{subsubsec:syntactic}
For syntactic information identification, sentences primarily consist of subjects, predicates, objects, and clauses. \textit{Syntactic Head} can distinctly identify and label nominal subjects, direct objects, adjectival modifiers, and adverbial modifiers.
Some words in the original sentence may get split into multiple subwords because of the tokenizer (e.g., ``happiness'' might be split into ``happi'' and ``ness''). The \textit{Subword Merge Head} focuses on these subwords and merges them into one complete word.\citep{InformationFlow_24_arXiv_Meta,SubwordHead_19_ACL_Portugal}

Additionally, \citet{KnowledgeCircuit_24_arXiv_ZJU} proposed the \textit{Mover Head} cluster, which can be considered as ``argument parsers''. These heads often copy or transfer a sentence's important information (such as the subject's position) to the [END] position. \revise{The [END] position refers to the last token's position in the sentence being decoded by the LLM, and many studies indicate that summarizing contextual information at this position facilitates subsequent reasoning and next-token prediction.}
\textit{Name Mover Head} and \textit{Backup Name Mover Head} move the names in the text to the [END] position.
\textit{Letter Mover Head} extracts the first letters of certain words in the context and aggregates them at the [END] position.\citep{AcronymPredict_24_arXiv_Alicante}
Conversely, \textit{Negative Name Mover Head} prevents name information from being transferred to the [END] position.\citep{IOI_23_ICLR_Redwood,CopySupression_23_arXiv_Google}

\subsubsection{Semantic Information Identification} \label{subsubsec:semantic}
As for semantic information identification, \textit{Context Head} extracts information from the context that is related to the current task.\citep{KnowledgeConflict_24_arXiv_UCAS}
Further, \textit{Content Gatherer Head} ``moves'' tokens related to the correct answer to the [END] position, preparing to convert them into the corresponding option letter for output.\citep{CorrectLetterHead_23_arXiv_DeepMind,ColorObject_24_ICLR_BrownU}
The \textit{Sentiment Summarizer} proposed by \citet{Sentiment_23_arXiv_EleutherAI} can summarize adjectives and verbs that express sentiment in the context near the [SUM] position. \revise{The [SUM] position is located directly before the [END] position and enables subsequent heads to effectively read and reason.}

Capturing the message about relationships is also important for future reasoning. \textit{Semantic Induction Head} captures semantic relationships within sentences, such as part-whole, usage, and category-instance relationships.\citep{Semantic_24_arXiv_SJTU}
\textit{Subject Head} and \textit{Relation Head} focus on subject attributes and relation attributes, respectively, and then inject these attributes into the residual stream.\citep{FactualRecall_24_arXiv_Independent}

\subsection{Latent Reasoning (LR)} \label{subsec:LR}
The KR and ICI stages focus on gathering information, while Latent Reasoning (LR) is where all the collected information is synthesized and logical reasoning occurs. Whether in humans or LLMs, the LR stage is the core of problem-solving. Specifically, QK matrices of a head perform implicit reasoning based on information read from the residual stream, and then the reasoning results or signals are written back into the residual stream through OV matrices.

\subsubsection{In-context Learning} \label{subsubsec:in-context}
In-context Learning is one of the most widely discussed areas. \revise{Building on the work of \citet{ICLDisentagle_23_thesis}, it primarily includes two types: Task Recognition (TR) and Task Learning (TL). Both aim to infer the solution based on the context; however, they differ fundamentally in their reliance on pre-trained priors. TR leverages the prior knowledge of LLMs to interpret demonstrations. For instance, sentiment classification tasks often involve labels with clear semantic meanings, such as ``positive'' and ``negative'', which LLMs are likely to have internalized during pre-training. In contrast, TL requires the model to learn a novel mapping function between input-output pairs, where the examples and labels lack an inherent semantic connection.}

For \textbf{Task Recognition}: \textit{Summary Reader} can read the information summarized at the [SUM] position during the ICI stage and use this information to infer the corresponding sentiment label.\citep{Sentiment_23_arXiv_EleutherAI}
\citet{FunctionVector_24_ICLR_NEU} proposed that the output of certain mid-layer attention heads can combine into a \textit{Function Vector}. These heads abstract the core features and logical relationships of a task, based on the semantic information identified during ICI, and thereby trigger task execution.

For \textbf{Task Learning}, the essence of solving these problems is enabling LLMs to inductively find patterns.
\textit{Induction Heads} are among the most widely studied attention heads.\citep{InductionHeads_22_TCT_Anthropic,Markov_24_arXiv_Harvard,FSL_24_ICML_UCL,InductionHead_24_arXiv_UoA} \revise{They capture patterns such as ``… \textcolor{mypurple}{[A]}[B] … \textcolor{deepgreen}{[A]}'' where token [B] follows token \textcolor{mypurple}{[A]}, and predict that the next token of this sequence should be [B].}
\revise{Specifically, the Induction Head in the residual stream of the second \textcolor{deepgreen}{[A]} can access information from that of all preceding tokens. This mainly includes information about ``what the previous token is'' for each token, which is provided by the Previous Head. The Induction Head then matches this information with the information in the current residual stream, i.e., it matches the second \textcolor{deepgreen}{[A]} with the \textcolor{mypurple}{[A]} preceding [B], to perform further reasoning.}

Induction Head tends to strictly follow a pattern once identified and completes fill-in-the-blank reasoning. However, in most cases, the real problem will not be identical to the examples—just as a student's exam paper will not be exactly the same as their homework. To address this, \citet{WordClassification_24_arXiv_UoM} \revise{identified} the \textit{In-context Head}, whose QK matrix calculates the similarity between information at the [END] position and each label. The OV matrix then extracts label features and weights them according to the similarity scores to determine the final answer (take all labels into consideration rather than only one label).

\subsubsection{Effective Reasoning} \label{subsubsec:EffectiveReason}
Some studies have identified heads related to reasoning effectiveness. \textit{Truthfulness Head} \citep{ITI_23_NIPS_harvard, NL-ITI_24_arXiv-Samsung} and \textit{Accuracy Head} \citep{CrossLingual_24_SIGIR_UCAS,LoFiT_24_arXiv_UT} are heads highly correlated with the truthfulness and accuracy of answers. They help the model infer truthful and correct results in QA tasks, and modifying the model along their activation directions can enhance LLMs' reasoning abilities. Similarly, the \textit{Consistency Head} ensures the internal consistency of LLMs when asked the same question in different ways.\citep{SemanticConsistency_24_ACL_TJU}

However, not all heads positively impact reasoning. For example, \textit{Vulnerable Head} is overly sensitive to certain specific input forms, making it susceptible to irrelevant information and leading to incorrect results.\citep{VulnerableHead_24_arXiv_Alicante} During reasoning, it is advisable to minimize the influence of such heads.

\subsubsection{Task Specific Reasoning} \label{subsubsec:TaskSpecific}
Finally, some heads are specialized for specific tasks.
In MCQA tasks, \textit{Correct Letter Head} can complete the matching between the answer text and option letters in order to determine the final answer choice.\citep{CorrectLetterHead_23_arXiv_DeepMind}
When dealing with tasks related to sequential data, \textit{Iteration Head} can iteratively infer the next intermediate state based on the current state and input.\citep{IterationHead_24_arXiv_Meta}
For arithmetic problems, \textit{Successor Head} can perform increment operations on ordinal numbers.\citep{SuccessorHead_24_ICLR_Cambridge}

In tasks such as syllogistic reasoning and information extraction, the Inhibition Head (also referred to as the Suppression Head) can aggregate outputs from other heads and suppress certain information. For example, it can suppress a subject or a middle term in order to reduce their associated logit values after unembedding.\citep{IOI_23_ICLR_Redwood,HeadCooperation_24_arXiv_UoM}

These examples illustrate how various attention heads specialize in different aspects of reasoning, contributing to the overall problem-solving capabilities of LLMs.

\subsection{Expression Preparation (EP)} \label{subsec:EP}
\begin{table}[htbp]
\centering
\caption{\revise{Key Attention Heads in Expression Preparation (EP).}}
\label{tab:EP_Head}
\resizebox{\textwidth}{!}{%
\revise{
\begin{tabular}{@{}lllc@{}}
\toprule
\multicolumn{1}{c}{\textbf{Head Name}} & \multicolumn{1}{c}{\textbf{Input Feature}} & \multicolumn{1}{c}{\textbf{Output Feature}} & \textbf{Layer Distribution} \\ \midrule
Mixed Head                             & Outputs of Subject \& Relation Heads       & Integrated and concise final representation & Deep                        \\
Amplification Head                     & Correct answer signals                     & Amplified attention on correct tokens       & Deep                        \\
Correct Head                           & Hidden states of different options         & Focused attentions on final output tokens   & Deep                        \\
Coherence Head                         & Contextualized reasoning outputs           & Fluent and coherent text's tokens           & Middle / deep               \\
Faithfulness Head                      & Reasoning results and instructions         & Selected faithful contexts                  & Deep                        \\
 \bottomrule
\end{tabular}%
}
}
\end{table}
During the Expression Preparation (EP) stage, LLMs need to align their reasoning results with the content that needs to be expressed verbally. \revise{As shown in Table \ref{tab:EP_Head},} EP heads may first \textbf{aggregate information} from various stages.
\citet{FactualRecall_24_arXiv_Independent} proposed the \textit{Mixed Head}, which can linearly combine and aggregate information \revise{passed along} by heads from the ICI and LR stages (such as Subject Heads, Relation Heads, Induction Heads, etc.). \revise{The aggregated results are then written back into the residual stream and ultimately mapped onto the vocabulary logits via the unembedding layer.}

Some EP heads have a \textbf{signal amplification} function. Specifically, they read information about the context or reasoning results from the residual stream, then enhance the information that needs to be expressed as output, and write it back into the stream.
\textit{Amplification Head} \citep{CorrectLetterHead_23_arXiv_DeepMind} and \textit{Correct Head} \citep{CorrectHead_24_arXiv_Allen} amplify the signal of the correct choice letter in MCQA problems near the [END] position. This amplification ensures that after passing through the Unembedding layer and softmax calculation, the correct choice letter has the highest probability.

In addition to information aggregation and signal amplification, some EP heads are used to \textbf{align the model's reasoning results with the user's instructions}.
In multilingual tasks, the model may sometimes fail to respond in the target language desired by the user. \textit{Coherence Head} ensures linguistic consistency in the generated content.\citep{CrossLingual_24_SIGIR_UCAS} They help LLMs maintain consistency between the output language and the language of user's query when dealing with multilingual inputs.
\textit{Faithfulness Head} is strongly associated with the faithfulness of \revise{Chain-of-Thought (CoT), which refers to whether the model's generated response accurately reflects its internal reasoning process and behavior, i.e., the consistency between output and internal reasoning.\citep{FaithfulCoT_24_ICML_Harvard}} Enhancing the activation of these heads allows LLMs to better align their internal reasoning with the output, making the CoT results more robust and consistent.

However, for some simple tasks, LLMs might not require special EP heads to refine language expression. In this situation, the information written back into the residual stream during the ICI and LR stages may be directly suitable for output, i.e., skip the EP stage and select the token with the highest probability. 

\begin{figure}[htbp]
    \centering
    \includegraphics[width=0.75\linewidth]{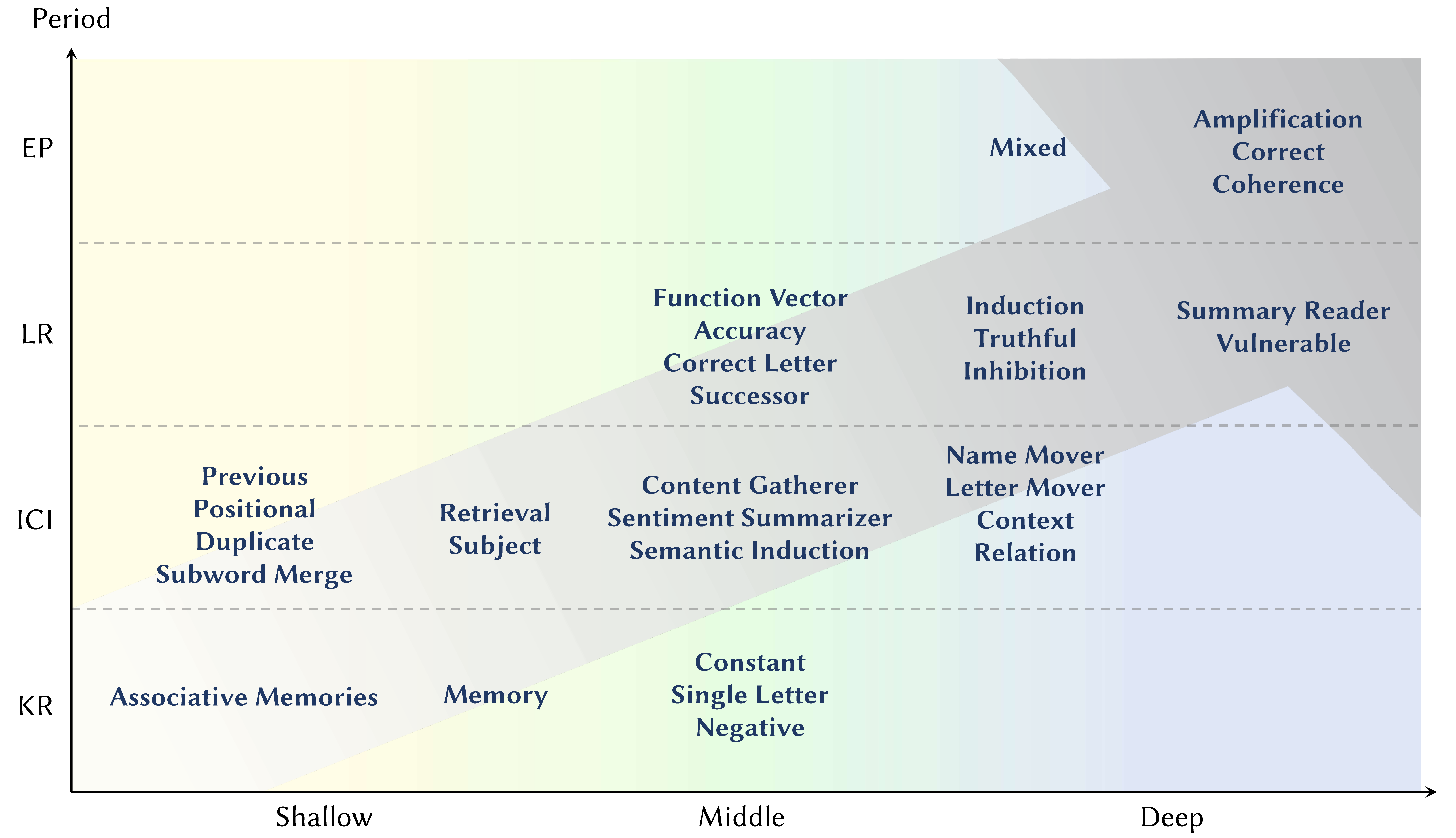}
    \caption{Diagram of the relationship between the stages where heads act and the layers they are in, as described from \nameref{subsec:KR} to \nameref{subsec:EP}.}
    \label{fig:4stage3layer}
\end{figure}

\subsection{How do attention heads work together?} \label{subsec:WorkTogether}
\revise{As illustrated in Figure~\ref{fig:4stage3layer}, if} we divide the layers of a LLM (e.g., GPT-2 Small) into three segments based on their order—shallow (e.g., layers 1-4), middle (e.g., layers 5-8), and deep (e.g., layers 9-12)—we can map the relationship between the stages where heads act and the layers they are in, according to the content above.
\revise{However, when combined with Figure~\ref{fig:four_steps}, this pattern reflects only the majority of cases; there are instances where LLMs return to the KR or ICI stage at deeper layers—for example, in the MCQA and IOI cases discussed below.}

\revise{To gain a enhanced understanding of the relationships between these heads, researchers have investigated} the \textbf{potential semantic meanings} embedded in the query vector $\mathbf{q}_{\ell, j}^h=\mathbf{Q}_{\ell}^h[:, j]$ and key vector $\mathbf{k}_{\ell, j}^h=\mathbf{K}_{\ell}^h[:, j]$.\citep{CorrectLetterHead_23_arXiv_DeepMind,ColorObject_24_ICLR_BrownU}
\revise{For example, when solving a MCQA problem, the model first infers the correct answer in text form. It then needs to map this text to the corresponding option letter based on the list of choices. At this point, during the ICI stage, the Content Gatherer Head moves the tokens of the inferred answer text to the [END] position.} Then, in the LR stage, the Correct Letter Head uses the information passed by the Content Gatherer Head to identify the correct option. The query vector in this context effectively asks, ``Are you the correct label?'' while recalling the gathered correct answer text. The key vector represents, ``I'm choice [A/B/C/D], with the corresponding text [...]''. After matching the right key vector to the query vector, we can get the correct answer choice.

\revise{Consider the Parity Problem, which involves determining whether the sum of an input sequence \colorbox{inputseq}{\mystrut $a_{1:t}$}, consisting of only $0$s and $1$s, is odd or even. Let parity state sequence \colorbox{parityseq}{\mystrut $s_i$} denote the parity (odd or even) of the sum of the first $i$ digits in the sequence, as defined in Equation~\ref{equ:parity}. For example, given the input sequence \colorbox{inputseq}{\mystrut $a_{1:6} = 001011$}, the corresponding parity state sequence is \colorbox{parityseq}{\mystrut $s_{1:6} = eeooeo$}. When querying the LLM with the prompt ``\colorbox{inputseq}{\mystrut $a_{1:t}$} [EOI] \colorbox{parityseq}{\mystrut $s_{1:t-1}$} [END]'', where [EOI] represents the End-Of-Input token, the expected response is the final parity state $s_t$.}

\revise{Under these settings,} during the ICI stage, a Mover Head transmits information from the [EOI] position to the [END] position.
In the LR stage, an Iteration Head first reads the [EOI]'s position index from [END] and uses its query vector to ask, ``Are you position $t$?'' The key vector for each token responds, ``I'm position $t^{'}$.'' This querying process identifies the last digit in the input sequence, which, combined with $s_{t-1}$, allows the model to calculate $s_{t}$.
\begin{equation}
\label{equ:parity}
\revise{
s_i = 
\left\{
\begin{aligned}
    & e, \quad \text{if } \sum_{k=1}^{i} a_k \text{ mod } 2 = 0, \\
    & o, \quad \text{if } \sum_{k=1}^{i} a_k \text{ mod }  2 = 1.
\end{aligned}
\right.
}
\end{equation}

\begin{figure}[htbp]
    \centering
    \includegraphics[width=0.85\linewidth]{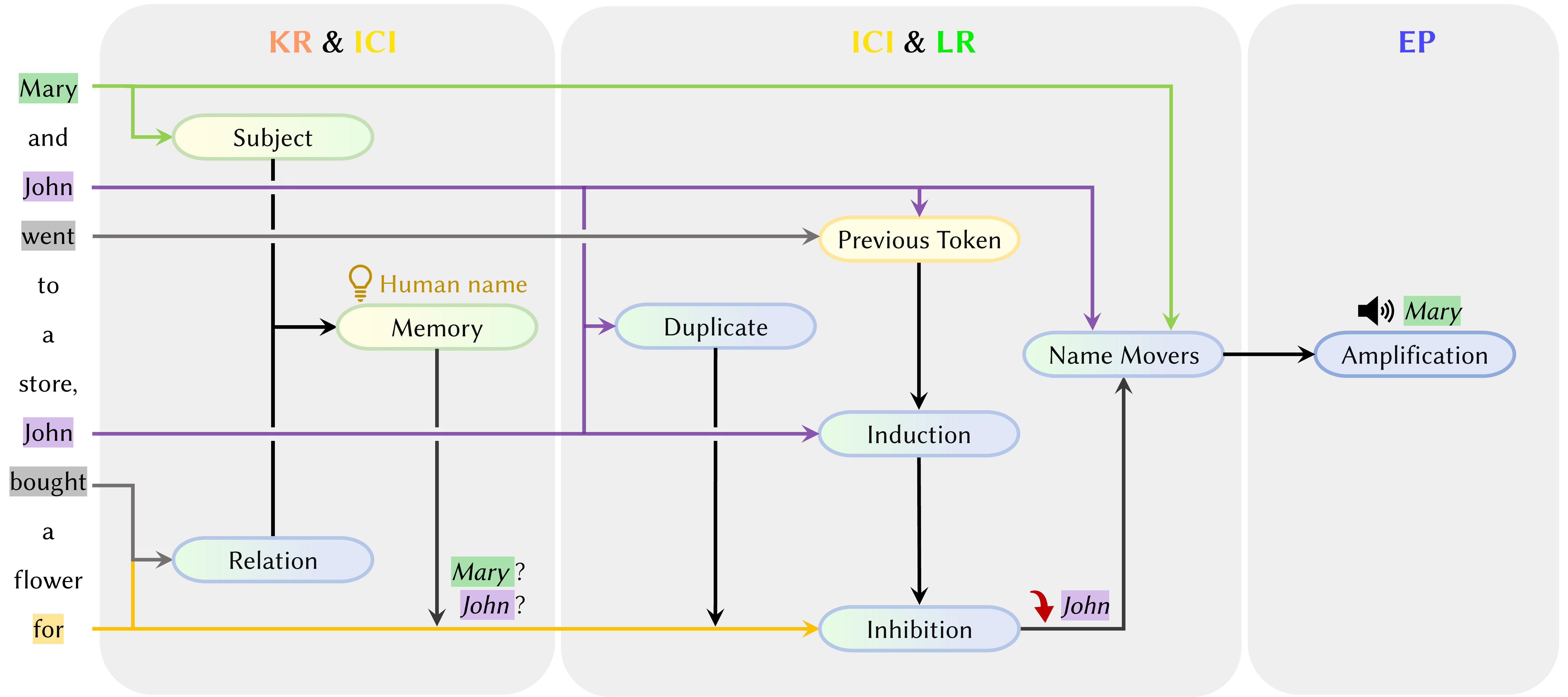}
    \caption{\revise{Schematic diagram of the collaborative mechanism of different attention heads in IOI task.\citep{IOI_23_ICLR_Redwood} Each oval represents a specific attention head, and the color indicates the depth of the layer where the head is located. These colors are aligned with those in Figure~\ref{fig:ResidualStream} and Figure~\ref{fig:4stage3layer}.}}
    \label{fig:IOIexample}
\end{figure}

Further research has explored integrating multiple special attention heads into a \textbf{cohesive working mechanism}.\citep{IOI_23_ICLR_Redwood,ColorObject_24_ICLR_BrownU,MechanCompet_24_ACL_ETH,HeadCooperation_24_arXiv_UoM} \revise{\citet{IOI_23_ICLR_Redwood}, \citet{ColorObject_24_ICLR_BrownU}, and \citet{HeadCooperation_24_arXiv_UoM} have independently identified the collaborative mechanisms of attention heads, such as mover heads, induction heads, and inhibition heads, in different task scenarios, namely Object Identification and Syllogistic Reasoning. Their studies, all conducted on the GPT-2 model,\citep{GPT2_MODEL} have yielded remarkably similar conclusions regarding the information transfer patterns among several key attention heads. Here we take the IOI (Indirect Object Identification) task, which tests the model's ability to deduce the indirect object in a sentence, as an example. Figure~\ref{fig:IOIexample} outlines the main collaboration process.}
\begin{enumerate}
    \item In the KR stage, the Subject Head and the Relation Head focus on ``Mary'' and ``bought flowers for'', respectively, triggering the model to recall that the answer should be a human name.\citep{FactualRecall_24_arXiv_Independent}

    \item Then in the ICI stage, the Duplicate Head identifies that ``John'' appears multiple times, while the Name Mover Head focuses on both ``John'' and ``Mary'' and moves them to the [END] position.

    \item During the iterative stages of ICI and LR, the Previous Head and the Induction Head work together to attend to ``John''. All this information is passed to the Inhibition Head, thereby suppressing the logits value of ``John''.
    \item Finally in the stage of EP, the Amplification Head boosts the logits value for ``Mary''.
\end{enumerate}

\revise{In summary, attention heads in LLMs work collaboratively across stages like KR, ICI, LR, and EP. This structured cooperation enables the model to solve complex tasks by effectively aligning and propagating relevant information through layers, further reflecting similarities between the working mechanisms of attention heads and the human brain.}

\section{Unveiling the discovery of attention heads} \label{sec:DiscoveryExp}
\begin{figure}[htbp]
    \centering
    \includegraphics[width=\linewidth]{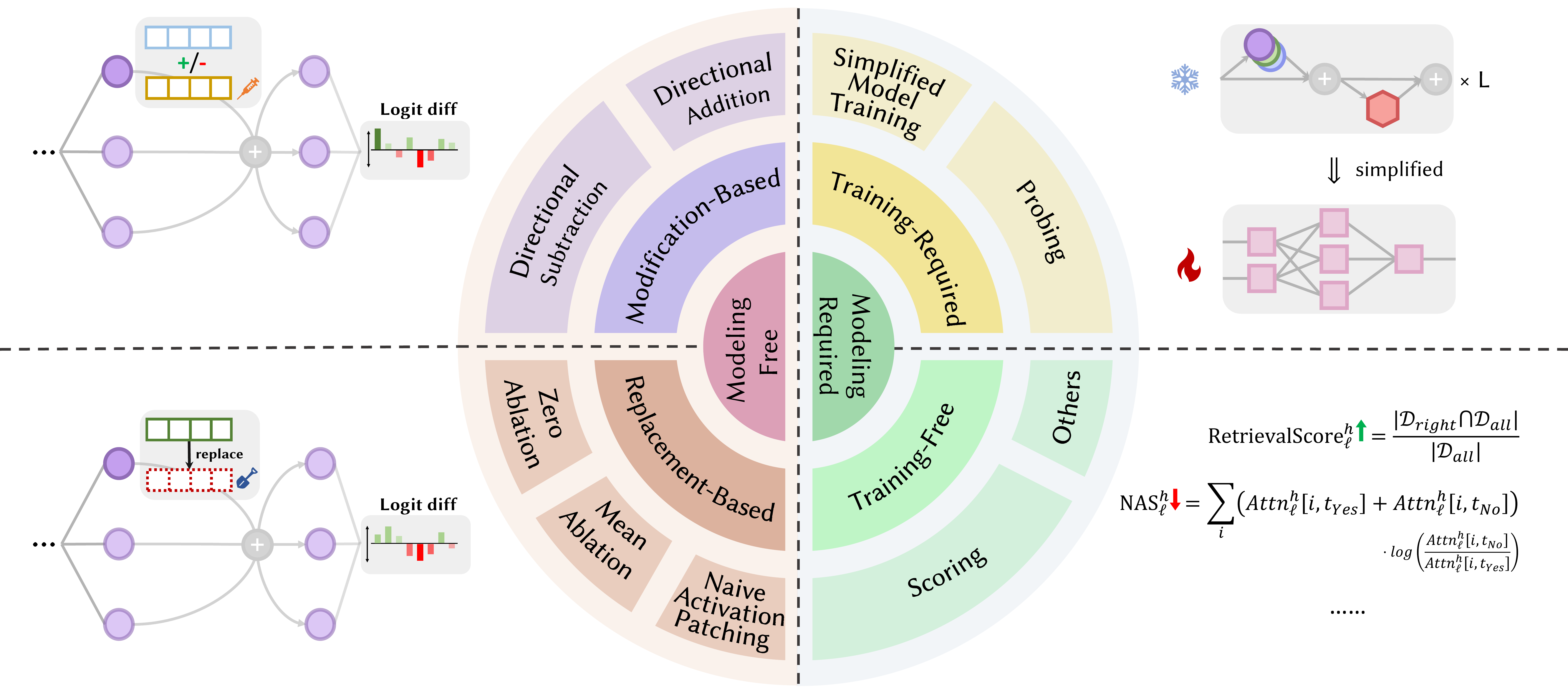}
    \caption{Pie chart of methods for exploring special attention heads and diagram of various methods.}
    \label{fig:piechart}
\end{figure}

How can we uncover the specific functions of those special heads mentioned in \nameref{sec:HeadOverview}? In this section, we will unveil the discovery methods. Current research primarily employs experimental methods to validate the working mechanisms of those heads. We categorize the mainstream experimental approaches into two types based on whether they require the construction of new models: Modeling-Free and Modeling-Required. The classification scheme and method examples are shown in Figure~\ref{fig:piechart}.

\subsection{Modeling-Free} \label{subsec:ModelFree}
\begin{table}[htbp]
\centering
\caption{Brief summarization of Modeling-Free methods.}
\label{tab:ModelingFree}
\resizebox{\textwidth}{!}{%
\begin{tabular}{@{}clll@{}}
\toprule
\textbf{Type}                                & \multicolumn{1}{c}{\textbf{Specific Method}} & \multicolumn{1}{c}{\textbf{Core Operation}}                                                                                                           & \multicolumn{1}{c}{\textbf{Representative Works}}                                                                                                                                                                                       \\ \midrule
\multirow{2}{*}{Modification-Based} & Directional Addition                & Adding extra information to a specific component's latent state                                                                              & \citet{Sentiment_23_arXiv_EleutherAI}, \citet{NegativeHead_24_arXiv_SNU}, \citet{ActivationAddition_23_arXiv_DeepMind}                                                                                                         \\ \cmidrule(l){2-4} 
                                    & Directional Subtraction             & Subtracting part of information from a specific component's latent state                                                                     & \citet{Sentiment_23_arXiv_EleutherAI}, \citet{DII_24_PMLR_Stanford}                                                                                                                                                                                          \\ \midrule
\multirow{6}{*}{Replacement-Based}  & Zero Ablation                       & The component's latent state is replace with zero vectors                                                                                    & \begin{tabular}[c]{@{}l@{}}\citet{IOI_23_ICLR_Redwood}, \citet{WordClassification_24_arXiv_UoM}, \citet{KnowledgeConflict_24_arXiv_UCAS},\\ \citet{KnowledgeCircuit_24_arXiv_ZJU}, \citet{ContextMix_23_EACL_Tilburg}\end{tabular}                                 \\ \cmidrule(l){2-4} 
                                    & Mean Ablation                       & \begin{tabular}[c]{@{}l@{}}The component's latent state is replace with the mean state across all \\ samples passing through it\end{tabular} & \begin{tabular}[c]{@{}l@{}}\citet{CopySupression_23_arXiv_Google}, \citet{IOI_23_ICLR_Redwood}, \citet{HeadCooperation_24_arXiv_UoM},\\\citet{Greater_23_NIPS_UoA}\end{tabular}
                                    
                                    \\ \cmidrule(l){2-4} 
                                    & Naïve Activation Patching           & \begin{tabular}[c]{@{}l@{}}The component's activation is replaced with corresponding activation \\ run by a corrupted prompt\end{tabular}     & \begin{tabular}[c]{@{}l@{}}\citet{ColorObject_24_ICLR_BrownU}, \citet{FunctionVector_24_ICLR_NEU}, \citet{IOI_23_ICLR_Redwood},\\ \citet{CorrectLetterHead_23_arXiv_DeepMind}, \citet{CorrectHead_24_arXiv_Allen}\end{tabular} \\ \bottomrule
\end{tabular}%
}
\end{table}

Modeling-Free methods do not require setting up new models, making them widely applicable in interpretability research. These methods typically involve altering a latent state computed during the LLMs' reasoning process and then using Logit Lens to map the intermediate results to token logits or probabilities. By calculating the logit (or probability) difference, researchers can infer the impact of the change. Modeling-Free methods primarily include Activation Patching and Ablation Study. However, due to the frequent interchange of these terms in the literature, a new perspective is required to distinguish them. This paper further divides these methods into Modification-Based and Replacement-Based Methods based on how the latent state representation is altered, as summarized in Table~\ref{tab:ModelingFree}.

\textbf{Modification-Based methods} involve altering the values of a specific latent state while retaining some of the original information, under the hypothesis that concepts are encoded as linear directions in the representation space.\citep{LinearRepresent_24_PMLR_UChicago}
Directional Addition retains part of the information in the original state and then directionally adds some additional information.

For instance, \citet{Sentiment_23_arXiv_EleutherAI} input texts containing positive and negative sentiments into LLMs, obtaining positive and negative representations from the latent state. The difference between these two representations can be seen as a sentiment direction in the latent space. By adding this sentiment direction vector to the activation of the attention head, the effect on the output can be analyzed to determine whether the head has the ability to summarize sentiment.
Similarly, \citet{MechanCompet_24_ACL_ETH} explored the competitive relationships between different mechanisms. They directionally amplified the attention score of one token towards another, allowing the latent representation to include more information about that token.

Conversely, Directional Subtraction retains part of the original state information while directionally removing some of it.\citep{SurveyCTG_24_arXiv_RUC} This method can be used to investigate whether removing specific information from a latent state affects the model's output in a significant way, thereby revealing whether certain attention heads can back up or fix the deleted information.

\textbf{Replacement-Based methods}, in contrast to Modification-Based methods, discard all information in a specific latent state and replace it with other values.
Zero Ablation and Mean Ablation replace the original latent state with zero values or the mean value of latent states across all samples from a dataset, respectively. This can logically ``eliminate'' the head or cause it to lose its special function, allowing researchers to assess its importance.

Naive Activation Patching is the traditional patching method. It involves using a latent state obtained from a corrupted prompt to replace the original latent state at the corresponding position. For example, consider the original prompt ``John and Mary went to the store.'' Replacing ``Mary'' with ``Alice'' results in a corrupted prompt. By systematically replacing the latent state obtained under the original prompt with the one obtained under the corrupted prompt across each head, researchers can preliminarily determine which head has the ability to focus on names based on the magnitude of the impact.\citep{ActivationPatching_24_arXiv_Google, ActivationPatching_24_ICLR_Google}
Alternatively, we can also replace the latent state obtained from the corrupted run with the original one. By doing so, we can observe how the head's behavior shifts back towards the performance on the original prompt.

\subsection{Modeling-Required} \label{subsec:ModelRequired}
\begin{table}[htbp]
\centering
\caption{Brief summarization of Modeling-Required methods.}
\label{tab:ModelingRequired}
\resizebox{\textwidth}{!}{%
\begin{tabular}{@{}clll@{}}
\toprule
\textbf{Type}                      & \multicolumn{1}{c}{\textbf{Specific Method}} & \multicolumn{1}{c}{\textbf{Core Operation}}                                                                                                       & \multicolumn{1}{c}{\textbf{Representative Works}}                                                                                                                                                                               \\ \midrule
\multirow{2}{*}{Training-Required} & Probing                                      & \begin{tabular}[c]{@{}l@{}}Train a classifier to distinguish heads with different func-\\ tions using activation values\end{tabular}              & \begin{tabular}[c]{@{}l@{}}\citet{ITI_23_NIPS_harvard}, \citet{NL-ITI_24_arXiv-Samsung}, \citet{SuccessorHead_24_ICLR_Cambridge},\\ \citet{CrossLingual_24_SIGIR_UCAS}, \citet{SemanticConsistency_24_ACL_TJU}, \citet{ConceptDepth_24_arXiv_Rutgers}\end{tabular}                                             \\ \cmidrule(l){2-4} 
                                   & Simplified Model Training                    & \begin{tabular}[c]{@{}l@{}}Train an approximate simplified model (e.g., a two-layer \\ Transformer or an attention-only model)\end{tabular}                                  & \begin{tabular}[c]{@{}l@{}}\citet{Markov_24_arXiv_Harvard}, \citet{IterationHead_24_arXiv_Meta}, \citet{InductionHead_24_ICLR_Princeton},\\\citet{MathFrame_21_TCT_Anthropic}  \end{tabular}                                                                                                                      \\ \midrule
\multirow{2}{*}{Training-Free}     & Scoring                                      & \begin{tabular}[c]{@{}l@{}}Calculate the score that reflects the relationship between \\ the component's attributes and LLM features\end{tabular} & \begin{tabular}[c]{@{}l@{}}\citet{KnowledgeConflict_24_arXiv_UCAS}, \citet{RetrievalHead_24_arXiv_PKU}, \citet{InductionHead_24_arXiv_UoA}, \citet{NegativeHead_24_arXiv_SNU},\\ \citet{HumanMemory_24_arXiv_UCSD}\end{tabular} \\ \cmidrule(l){2-4} 
                                   & Others                                       & New methods that have not yet been widely adopted                                                                                                 & \citet{InformationFlow_24_arXiv_Meta}, \citet{ACDC_23_NIPS_UCL}                                                                                                                                                                                           \\ \bottomrule
\end{tabular}%
}
\end{table}
Modeling-Required methods involve explicitly constructing models to delve deeper into the functions of specific heads. Based on whether the newly constructed models require training, we further categorize Modeling-Required methods into Training-Required and Training-Free methods, as summarized in Table~\ref{tab:ModelingRequired}.

\textbf{Training-Required methods} necessitate training the newly established models to explore mechanisms.
Probing is a common training-based method. This approach extracts activation values from different heads as features and categorizes heads into different classes as labels. A classifier is then trained on this data to learn the relationship between the activation patterns and the head's function. Subsequently, the trained classifier can serve as a probe to detect which heads within the LLMs possess which functions.\citep{SemanticConsistency_24_ACL_TJU, ITI_23_NIPS_harvard}

Another approach involves training a simplified transformer model on a clean dataset for a specific task. Researchers investigate whether the heads in this simplified model exhibit certain functionalities, which can then be extrapolated whether similar heads in the original model possess the same capabilities. This method reduces computational costs during training and analysis, while the constructed model remains simple and highly controllable.\citep{IterationHead_24_arXiv_Meta}

\textbf{Training-Free methods} primarily involve designing scores that reflect specific phenomena. These scores can be viewed as mathematical models that construct an intrinsic relationship between the attributes of components and certain model characteristics or behaviors.
For instance, when investigating Retrieval Heads, \citet{RetrievalHead_24_arXiv_PKU} defined a Retrieval Score. This score represents the frequency with which a head assigns the highest attention score to the token it aims to retrieve across a sample set, as shown in Equation~\ref{equ:RetrievalScore}. A high Retrieval Score indicates that the head possesses a strong ``Needle in a Haystack'' ability.

Similarly, when exploring Negative Heads, \citet{NegativeHead_24_arXiv_SNU} introduced the Negative Attention Score (NAS), as shown in Equation~\ref{equ:NAS}. Here, $i$ denotes the $i$-th token in the input prompt, and $t_{Yes}$ and $t_{No}$ represent the positions of ``Yes'' and ``No'' in the prompt, respectively. A high NAS suggests that the head focuses more on negative tokens during decision-making, making it prone to generating negative signals.
\begin{equation} \label{equ:RetrievalScore}
\text{RetrievalScore}_{\ell}^{h} = \frac{|\mathcal{D}_{right} \cap \mathcal{D}_{all}|}{|\mathcal{D}_{all}|}
\end{equation}
\begin{equation} \label{equ:NAS}
\text{NAS}_{\ell}^{h} = \sum_{i}{\left(Attn_{\ell}^{h}\left[i, t_{Yes}\right] + Attn_{\ell}^{h}\left[i, t_{No}\right]\right) \cdot \log{\left(\frac{Attn_{\ell}^{h}\left[i, t_{No}\right]}{Attn_{\ell}^{h}\left[i, t_{Yes}\right]}\right)}}
\end{equation}

In addition to scoring, researchers have proposed other novel training-free modeling methods.
\citet{InformationFlow_24_arXiv_Meta} introduced the concept of an Information Flow Graph (IFG), where nodes represent tokens and edges represent information transfer between tokens via attention heads or FFNs. By calculating and filtering the importance of each edge to the node it points to, key edges can be selected to form a subgraph. This subgraph can then be viewed as the primary internal mechanism through which LLMs perform reasoning.

\section{Evaluation} \label{sec:Evaluation}
This section summarizes the benchmarks and datasets used in the interpretability research of attention heads. Based on the different evaluation goals during the mechanism exploration process, we categorize them into two types: Mechanism Exploration Evaluation and Common Evaluation. The former is designed to evaluate the working mechanisms of specific attention heads, while the latter assesses whether enhancing or suppressing the functions of certain special heads can improve the overall performance of LLMs.
\begin{table}[htbp]
    \centering
    \caption{Selected benchmarks for mechanism exploration evaluation.}
    \label{tab:MechanismEval}
    \resizebox{\textwidth}{!}{%
    \begin{tabular}{@{}llllc@{}}
    \toprule
    \textbf{Benchmark}                                    & \textbf{Type}         & \textbf{Main Task}                                 & \textbf{Source} & \textbf{Release Date} \\ \midrule
    LRE~\citep{LRE_MIT}                                   & Knowledge recalling   & Infer object entities given subject-entity prompts & MIT             & 2023.09               \\
    ToyMovieReview~\citep{Sentiment_23_arXiv_EleutherAI}  & Sentiment analysis    & Infer positive/negative sentiment                  & EleutherAI      & 2023.10               \\
    ToyMoodStory~\citep{Sentiment_23_arXiv_EleutherAI}    & Sentiment analysis    & Infer positive/negative sentiment                  & EleutherAI      & 2023.10               \\
    FV-Caplitalize~\citep{FunctionVector_24_ICLR_NEU}     & Token-level reasoning & Infer the capital letter given some words          & NEU             & 2023.10               \\
    ICL-MC~\citep{Markov_24_arXiv_Harvard}                & Token-level reasoning & Infer next state based on in-context               & Harvard         & 2024.02               \\
    Succession~\citep{SuccessorHead_24_ICLR_Cambridge}    & Arithmetic reasoning  & Infer next number in a incremental sequence        & Cambridge       & 2023.12               \\
    Iteration-Synthetic~\citep{IterationHead_24_arXiv_Meta} & Arithmetic reasoning  & Infer the next state of an iterative process       & Meta            & 2024.06               \\
    Omniglot~\citep{Omniglot_15_NYU}                      & Word-level reasoning  & Infer label from few samples                       & NYU             & 2015.12               \\
    IOI~\citep{IOI_23_ICLR_Redwood}                       & Word-level reasoning  & Infer the indirect object                          & UCB             & 2022.11               \\
    Colored Object~\citep{ColorObject_24_ICLR_BrownU}     & Word-level reasoning  & Infer the correct color of a material              & Brown U         & 2023.10               \\
    World-Capital~\citep{KnowledgeConflict_24_arXiv_UCAS} & Word-level reasoning  & Infer the capital city given a country             & UCAS            & 2024.02               \\ \bottomrule
    \end{tabular}%
    }
\end{table}

\subsection{Mechanism exploration evaluation}
To delve deeper into the internal reasoning paths of LLMs, many researchers have synthesized new datasets based on existing benchmarks. The primary feature of these datasets is the simplification of problem difficulty, with elements unrelated to interpretability, such as problem length and query format, being standardized. As shown in Table~\ref{tab:MechanismEval}, these datasets essentially evaluate the model's knowledge reasoning and knowledge recalling capabilities, but they simplify the answers from a paragraph-level to a token-level.

Take exploring sentiment-related heads as an example, \citet{Sentiment_23_arXiv_EleutherAI} created the ToyMovieReview and ToyMoodStory datasets, with specific prompt templates illustrated in Figure~\ref{fig:SentimentPrompt}. Using these datasets, researchers employed sampling methods to calculate the activation differences of each head for positive and negative sentiments. This allowed them to recognize heads with significant differences as potential candidates for the role of Sentiment Summarizers.
\begin{figure}[!ht]
\centering

\begin{tcolorbox}[colback=blue!5!white,colframe=blue!75!black,fontupper=\footnotesize,fonttitle=\scriptsize]
\small
\textbf{ToyMovieReview}: \\
I thought this movie was \textcolor{orange}{ADJECTIVE}, I \textcolor{red}{VERB}ed it. Conclusion: This movie is \_\_\_\_\_.\\
\\
\textbf{ToyMoodStory}: \\
\textcolor{purple}{$\text{NAME}_1$} \textcolor{red}{$\text{VERB}_1$} parties, and \textcolor{red}{$\text{VERB}_2$} them whenever possible.\\
\textcolor{purple}{$\text{NAME}_2$} \textcolor{red}{$\text{VERB}_3$} parties, and \textcolor{red}{$\text{VERB}_4$} them whenever possible.\\
One day, they were invited to a grand gala. [\textcolor{purple}{$\text{NAME}_1$} or \textcolor{purple}{$\text{NAME}_2$}] feels very \_\_\_\_\_.
\end{tcolorbox}
\caption{Prompt template for ToyMovieReview and ToyMoodStory dataset. For example, \textcolor{orange}{ADJECTIVE} could be ``fantastic'' / ``horrible'', \textcolor{red}{VERB} could be ``like'' / ``dislike''.}

\label{fig:SentimentPrompt}
\end{figure}

\subsection{Common evaluation}
\begin{table}[htbp]
\centering
\caption{Selected benchmarks for common evaluation.}
\label{tab:CommonEval}
\resizebox{\textwidth}{!}{%
\begin{tabular}{@{}llllc@{}}
\toprule
\textbf{Benchmark}                                                                 & \textbf{Type}          & \textbf{Main Task}                        & \textbf{Source} & \textbf{Release Date} \\ \midrule
MMLU~\citep{MMLU_UCB}                                                              & Knowledge reasoning    & Solve problems with widespread knowledge  & UCB             & 2020.09               \\
TruthfulQA~\citep{TruthfulQA_Oxford}                                               & Knowledge reasoning    & Answer questions that span 38 categories  & Oxford          & 2021.09               \\
LogiQA~\citep{LogiQA_20_IJCAI_FDU}                                                 & Logic resoning         & Deduce the answer of logical problems     & FDU             & 2020.07               \\
MQuAKE~\citep{MQuAKE_23_EMNLP_Princeton}                                           & Logic resoning         & Deduce the answer via multi-hop reasoning & Princeton       & 2023.05               \\
SST/SST2~\citep{SST_Stanford}                                                      & Sentiment analysis     & Infer positive/negative sentiment         & Standford       & 2013.10               \\
ETHOS~\citep{ETHOS_20_arXiv_Greece}                                                & Sentiment analysis     & Detect hate speech in online comments     & AUT             & 2020.06               \\
\href{https://github.com/gkamradt/LLMTest_NeedleInAHaystack}{Needle-in-a-Haystack} & Long context retrieval & Retrieve content from long context        & Github          & 2023.11               \\
AG News~\citep{AGNews_NYU}                                                         & Text comprehension     & Infer the category of news                & NYU             & 2015.02               \\
TriviaQA~\citep{TriviaQA_UoW}                                                      & Text comprehension     & Answer questions based on documents       & UoW             & 2017.05               \\
AGENDA~\citep{AGENDA_UoW}                                                          & Text comprehension     & Generate the abstract of a passage        & UoW             & 2019.04               \\ \bottomrule
\end{tabular}%
}
\end{table}

The exploration of attention head mechanisms is ultimately aimed at improving the comprehensive capabilities of LLMs. Many researchers, upon identifying a head with a specific function, have attempted to modify that type of head—such as by enhancing or diminishing its activation—to observe whether the LLMs' responses become more accurate and useful. We classify these Common Evaluation Benchmarks based on their evaluation focus, as shown in Table~\ref{tab:CommonEval}. The special attention heads discussed in this paper are closely related to improving LLMs' abilities in five key areas: knowledge reasoning, logic reasoning, sentiment analysis, long context retrieval, and text comprehension.

\section{Additional topics} \label{sec:other_tasks}
In this section, we summarize various works related to the LLMs interpretability. Although these works may not recognize new special heads as discussed in \nameref{sec:HeadOverview}, they delve into the underlying mechanisms of LLMs from other perspectives. We will elaborate on these studies under two categories: FFN Interpretability and Machine Psychology.

\subsection{FFN interpretability} \label{subsec:OtherComponent}
As discussed in \nameref{sec:background}, apart from attention heads, FFNs also plays a significant role in the LLMs reasoning process. This section primarily summarizes research focused on the mechanisms of FFNs and the collaborative interactions between attention heads and FFNs.

One of the primary functions of FFNs is to store knowledge acquired during the pre-training phase. 
\citet{MLPKnowledge} proposed that factual knowledge stored within the model is often concentrated in a few neurons of the MLP, reflecting the sparsity of the model.\citep{DeadNeuron_24_ACL_Meta}
\citet{TransformerKVpair} observed that the neurons in the FFN of GPT models can be likened to key-value pairs, where specific keys can retrieve corresponding values, i.e., knowledge.
\citet{FactualRecall_24_arXiv_RUC} discovered a hierarchical storage of knowledge within the model's FFN, with lower layers storing syntactic and semantic information, and higher layers storing more concrete factual content.

FFNs effectively complement the capabilities of attention heads across the four stages described in \nameref{sec:HeadOverview}. The collaboration between FFNs and attention heads enhances the overall capabilities of LLMs.
\citet{FacutalRecall_23_EMNLP_Google} proposed that attention heads and FFNs can work together to enrich the representation of a subject and then extract its related attributes, thus facilitating factual information retrieval during the Knowledge Recall (KR) stage.
\citet{MLPLocalUpdate_23_EMNLP_ETH} found that, unlike attention heads, which focus on global information and perform aggregation, FFNs focus only on a single representation and perform local updates. This complementary functionality allows them to explore textual information both in breadth (attention heads) and depth (FFNs).

In summary, each component of LLMs plays a crucial role in the reasoning process. The individual contributions of these components, combined with their interactions, accomplish the entire process from Knowledge Recalling to Expression.

\subsection{Machine Psychology} \label{subsec:MachinePsychology}
Current research on the LLMs interpretability often draws parallels between the reasoning processes of these models and human thinking. This suggests the need for a more unified framework that connects LLMs with human cognition. The concept of Machine Psychology has emerged to fill this gap,\citep{MachinePsychologyOrigin} exploring the cognitive activities of AI through psychological paradigms.

Recently, \citet{MachinePsychology_Hagendorff} and \citet{MachinePsychology_Johansson} have proposed different approaches to studying machine psychology.
Hagendorff's work focuses on using psychological methods to identify new abilities in LLMs, such as heuristics and biases, social interactions, language understanding, and learning. His research suggests that LLMs display human-like cognitive patterns, which can be analyzed to improve AI interpretability and performance.

Johansson's framework integrates principles of operant conditioning \citep{OperantConditioning} with AI systems, emphasizing adaptability and learning from environmental interactions. This approach aims to bridge gaps in AGI research by combining insights from psychology, cognitive science, and neuroscience.

Overall, Machine Psychology provides a new perspective for analyzing LLMs. Psychological experiments and behavioral analyses may lead to new discoveries about these models. As LLMs are increasingly applied across various domains of society, understanding their behavior through a psychological lens becomes increasingly important, which offers valuable insights for developing more intelligent AI systems.

\section{Discussion} \label{sec:Discussion}
\subsection{\revise{Limitations in existing studies}}
\revise{Although substantial progress has been made in uncovering the internal mechanisms of LLMs, several key limitations persist in existing research. These can be summarized as follows:}
\begin{itemize}
    \item \revise{\textbf{Lack of task generalizability}.} Current research primarily explores simple application scenarios that are limited to specific types of tasks. For example, \citet{IOI_23_ICLR_Redwood} and \citet{ColorObject_24_ICLR_BrownU} have identified reasoning circuits in LLMs through tasks such as the IOI task and the Color Object Task. However, these circuits have not been validated across other tasks, making it challenging to determine whether these mechanisms are universally applicable.

    \item \revise{\textbf{Lack of Mechanism Transferability}. As shown in Figure~\ref{fig:head_taxnomomy}, many discovered special heads have only been explored within a few specific LLMs, or even on custom-built toy models. This raises a critical question: does a specialized head identified in one LLM exhibit the same functionality in another LLM? However, current research lacks investigations into the transferability of such mechanisms across different model series.}
    
    \item \revise{\textbf{Limited focus on multi-head collaboration}.} Most studies investigate the mechanisms of individual attention heads, with only a few researchers studying the collaborative relationships among multiple heads. Consequently, existing work lacks a comprehensive framework for understanding the coordinated functioning of all attention heads in LLMs and analogizing the human brains.

    \item \revise{\textbf{Absence of theoretical supports}.} Many studies propose hypotheses about circuits based on observed phenomena and validate these hypotheses through experiments. However, this approach cannot establish the theoretical soundness of the mechanisms, nor can it determine whether the observed mechanisms are merely coincidental.
\end{itemize}

\subsection{Future directions and challenges}
Building on the limitations discussed above and the content presented earlier, this paper outlines several potential research directions for the future:
\begin{itemize}
    \item \textbf{Exploring mechanisms in more complex tasks}. Investigate whether certain attention heads possess special functions in more complex tasks, such as open-ended question answering,\citep{XSUM_18_arXiv,NewsBench_24_ACL_Melbourne} math problems \citep{MATH_21_arXiv,GSM8K_21_arXiv} and tool-using tasks \citep{TEval_24_ACL_USTC}.
    
    \item \textbf{Mechanism's robustness against prompts}. Research has shown that current LLMs are highly sensitive to prompts, with slight changes potentially leading to opposite responses.\citep{xFinder_24_arXiv_IAAR} Future work could analyze this phenomenon through the lens of attention head mechanisms and propose solutions to mitigate this issue.
    
    \item \textbf{Developing new experimental methods}. Explore new experimental approaches, such as designing experiments to verify whether a particular mechanism is indivisible or whether it has universal applicability.
    
    \item \textbf{Building a comprehensive interpretability framework}. This framework should encompass both the independent and collaborative functioning mechanisms of most attention heads and other components.
    
    \item \textbf{Integrating Machine Psychology}. Incorporate insights from Machine Psychology to construct an internal mechanism framework for LLMs from an anthropomorphic perspective, understanding the gaps between current LLMs and human cognition and guiding targeted improvements.
\end{itemize}

\section{Limitation} \label{sec:Limitation}
Current research on the interpretability of LLMs’ attention heads is relatively scattered, primarily focusing on the functions of individual heads, \revise{while lacking a rigorous definition of the overall framework.} As a result, the categorization of attention head functions from the perspective of human cognitive behavior in this paper may not be perfectly orthogonal, potentially leading to some overlap between different stages.

\section*{Experimental procedures}
\subsection*{Resource availability}
\subsubsection*{Lead contact}
Additional information, questions, and requests should be directed to the lead contact, Dr. Zhiyu Li (\href{mailto:lizy@iaar.ac.cn}{lizy@iaar.ac.cn}).
\subsubsection*{Materials availability}
Not applicable, as no new unique reagents were generated.
\subsubsection*{Data and code availability}
Our reference list is available at GitHub (\href{https://github.com/IAAR-Shanghai/Awesome-Attention-Heads}{https://github.com/IAAR-Shanghai/Awesome-Attention-Heads}).


\section*{Author contributions}
Conceptualization, Z.Z., Y.W. and S.S.; planning, Z.Z.; investigation: Z.Z. and Y.W.; original draft, Z.Z. and Y.H.; visualization, Z.Z. and Y.H.; review \& editing, all authors; project administration, M.Y., B.T., F.X. and Z.L.; supervision, Z.L.

\section*{Declaration of interests}
The authors declare no competing interests.

\bibliography{bib/AttnHead,bib/Eval+Addi,bib/Exp,bib/Neuro,bib/Others,bib/Survey} 

\end{document}